\documentclass[times,twocolumn,final,authoryear]{elsarticle}

\usepackage{framed,multirow}

\makeatletter
\def\ps@pprintTitle{%
  \let\@oddhead\@empty
  \let\@evenhead\@empty
  \let\@oddfoot\@empty
  \let\@evenfoot\@oddfoot
}
\makeatother

\usepackage{amssymb}
\usepackage{latexsym}

\usepackage{url}
\usepackage{xcolor}
\definecolor{newcolor}{rgb}{.8,.349,.1}
\usepackage{graphicx}
\usepackage{booktabs}
\usepackage{amsmath}
\usepackage{amssymb}
\usepackage{tabularx}
\usepackage{amsmath}
\usepackage{arydshln}
\usepackage{hyperref}

\definecolor{ForestGreen}{RGB}{1,68,33}
\definecolor{LightBlue}{rgb}{0.68, 0.85, 0.9}

\usepackage{multirow}

\usepackage{amsmath}
\DeclareMathOperator*{\argmax}{argmax} % thin space, limits underneath in displays

\makeatletter
\def\Ddots{\mathinner{\mkern1mu\raise\p@
\vbox{\kern7\p@\hbox{.}}\mkern2mu
\raise4\p@\hbox{.}\mkern2mu\raise7\p@\hbox{.}\mkern1mu}}
\makeatother

\journal{Expert Systems with Applications}

\begin{document}

\ifpreprint
  \setcounter{page}{1}
\else
  \setcounter{page}{1}
\fi

\begin{frontmatter}

\title{Knowledge-driven Answer Generation for Conversational Search~\textsuperscript{*}}
\cortext[cor1]{Pre-print paper.}

%\author[1]{\corref{cor1}Mariana \snm{Leite}}
\author[1]{\corref{cor1}Mariana Leite}
\ead{me.leite@campus.fct.unl.pt}
%\author[1]{Rafael \snm{Ferreira}}
\author[1]{Rafael Ferreira}
\ead{rah.ferreira@campus.fct.unl.pt}
%\author[1]{David \snm{Semedo}}
\author[1]{David Semedo}
\ead{df.semedo@fct.unl.pt}
%\author[1]{João \snm{Magalhães}} 
\author[1]{João Magalhães} 
\ead{jmag@fct.unl.pt}

%Affiliation 1, Address, City and Postal Code, Country
\address[1]{Universidade NOVA de Lisboa, Portugal}

% \received{1 January 2021}
% \finalform{2 January 2021}
% \accepted{3 January 2021}
% \availableonline{4 January 2021}
% \communicated{J. Smith}

\begin{abstract}
The conversational search paradigm introduces a step change over the traditional search paradigm by allowing users to interact with search agents in a multi-turn and natural fashion. The conversation flows naturally and is usually centered around a target field of knowledge. In this work, we propose a knowledge-driven answer generation approach for open-domain conversational search, where a conversation-wide entities' knowledge graph is used to bias search-answer generation. First, a conversation-specific knowledge graph is extracted from the top passages retrieved with a Transformer-based re-ranker. The entities knowledge-graph is then used to bias a search-answer generator Transformer towards information rich and concise answers. This conversation specific bias is computed by identifying the most relevant passages according to the most salient entities of that particular conversation.
Experiments show that the proposed approach successfully exploits entities knowledge along the conversation, and outperforms a set of baselines on the search-answer generation task.
\end{abstract}

% \begin{keyword}
% \MSC 41A05\sep 41A10\sep 65D05\sep 65D17
% \KWD Keyword1\sep Keyword2\sep Keyword3
% %Conversational Search, Answer Generation,  Knowledge Graph,  Entities.

% %% MSC codes here, in the form: \MSC code \sep code
% %% or \MSC[2008] code \sep code (2000 is the default)
% \end{keyword}

\end{frontmatter}

%\linenumbers

\section{Introduction}
In conversational search systems, users can interact in a natural manner with search systems. These go beyond the traditional search task where, on a multi-turn session, users query the system until the information need is met, thus resembling the way humans interact with each other. Supporting this paradigm shift are the observations made in~\cite{evaluation_conversational_assistants}, which revealed that users are receptive to conversational systems, provided that they meet users' expectations with respect to information seeking.
%Conversational search systems are an emerging research topc, and the natural evolution of the traditional search paradigm, allowing for a more natural interaction between users and search systems. Building intelligent systems able to establish and develop meaningful conversations is one of the key goals of AI and the ultimate goal of natural language research~\cite{wizardofwikipedia}. 
%The interactions between a user and conversational systems have been studied in ~\cite{evaluation_conversational_assistants}, which showed that users are willing to utilise conversational assistants as long as their needs are met with success.
%Conversational search assistants still put a considerable burden on users that have to go through a list of documents, or passages, to find the information they need.
To this extent, two main challenges must be addressed~\cite{challenges_dialog_systems,evaluation_conversational_assistants}: a) keep track of the dialog context and b) generate informative yet concise search-answers. To address a), conversational search systems adopt a query rewriting-based approach~\cite{t5conversational,limited_supervision_query_rewrite}, which rewrites a conversational query in order to make it context-independent. Then, retrieval and re-ranking are performed to retrieve a set of relevant passages. Addressing b) requires going beyond passage retrieval, and generate a short search-answer, similar to what is accomplished in chit-chat dialogue systems~\cite{retrieval_based_generation_1,knowledge-aware-dialogue-generation,retrieval_based_generation_2}. We argue that knowledge from current and previous turns is still crucial to provide the user with the most informative answer, and should be used seamlessly in addressing challenges a) and b). Knowledge about entities has proved to be important for search tasks~\cite{entity_query_expansion,context_ranking_entities,entity_saliency_retrieval}. Over the conversation, the interactions between different entities are expected to implicitly encode the conversational context. Therefore, whether the query re-writer perfectly manages to do coreference resolution or not, the entities that are not mentioned in the current query, but from previous turns, also shape the conversation context.

In this paper we propose a knowledge-driven answer generation system for conversational search that is aware of the context of the conversation to generate abstractive and knowledge-driven responses. Specifically, the knowledge about entities' interactions across a conversation will be modelled and used to condition the generation of a single and short search-answer, based on the information comprised in top-retrieved passages. 
Hence, the core research hypothesis of this paper is that \textit{a conversational agent's most informative answer can be generated by considering the intersection of the rank of passages and its graph of entities}. In particular, we take a knowledge-driven approach to guide the generation of the answer. 
First, the framework is built on top of a solid conversational response-retrieval method that is on par with state-of-the-art results on conversational search~\cite{castoverview}. 
Second the answer generation is leveraged by a conversation-wide entities knowledge graph, and biased according to the relations with the entities present on the query and top passages. Therefore, combining an entities knowledge-base with a strong conversational passage ranking baseline, allows scoring already highly relevant individual passages, according to their entities relations. These can then be fed to an answer-generation Transformer~\cite{vaswani2017attention}, that will produce a knowledge enriched agent response, biased by the knowledge-base. This enables the creation of richer answers covering a wider range of information, i.e., more comprehensive answers.

Next, we discuss the related work. Section~\ref{sec_related_work} details the state-of-the-art conversational response retrieval method. Section~\ref{sec_answer_gen} proposes a knowledge-driven answer-generator. Evaluation and results discussion is presented in sections \ref{sec_evaluation}, \ref{sec_results}, and \ref{sec_discussion}, and concluding remarks in section~\ref{sec_conclusions}.

%Some Papers that may be useful
%https://arxiv.org/pdf/1910.13461.pdf 
%https://arxiv.org/pdf/1912.08777.pdf 
%https://link.springer.com/chapter/10.1007/978-3-030-45439-5_6 
%https://link.springer.com/chapter/10.1007/978-3-030-45442-5_23 
%https://link.springer.com/chapter/10.1007/978-3-030-45439-5_46 
%https://link.springer.com/chapter/10.1007/978-3-030-45442-5_22 
%https://link.springer.com/chapter/10.1007/978-3-030-45442-5_32 
%https://dl.acm.org/doi/10.1145/3397271.3401110 
%https://dl.acm.org/doi/10.1145/3397271.3401130 
%https://arxiv.org/pdf/2005.11364.pdf

% https://chauff.github.io/documents/publications/KDDConverse2020-Penha.pdf

\section{Related Work}
\label{sec_related_work}
\noindent\textbf{Open-domain conversational assistant.}
Research on interactive search systems started a long time ago, with the goal of developing artificial intelligent conversation search agents to aid users in a variety of search tasks in a natural manner~\cite{belkin1980anomalous,croft_newapproach_i3r,oddy_information_1977}. With recent developments on machine learning and deep neural networks, together with improvements on computational infrastructures, the field is once again highly active.
%Open-domain conversational search systems must account for the dialog context to provide a relevant passage. While research on interactive search systems has started long ago~\cite{oddy_information_1977,belkin1980anomalous,croft_newapproach_i3r}, the recent interest in having intelligent conversation assistants (e.g. Alexa, SIRI), has re-ignited this research field. 
Namely, very recently the TREC CAsT (Conversational Assistant Track)~\cite{castoverview} task introduced a multi-turn passage retrieval dataset, supporting research on conversational search systems.
Current state-of-the-art approaches~\cite{wizardofwikipedia,t5conversational,open-retrieval-qa-sigir2020,limited_supervision_query_rewrite} overcome the need for abundant labelled data, by training self-supervised neural models on large and wide (w.r.t. topic coverage) collections such as Wikipedia \cite{bertOriginal,roberta,yang2019xlnet}. This results in rich language models that can be applied to the several components of a conversational search agent pipeline, including addressing conversational context and passage re-ranking in each turn.
%leverage on large open-domain collections (e.g. Wikipedia) to learn rich language-models using self-supervised neural networks. 
%The applicability of these models is twofold: grasping the dialog context and passage re-ranking. 
%Towards addressing dialog context, query re-rewriting is one of the approaches adopted to obtain context-independent queries based on previous utterances~\cite{canYouUnpackIt,limited_supervision_query_rewrite}. 
Transformer models, pre-trained on large collections, have been used lately for both the task of query rewriting~\cite{t5conversational,limited_supervision_query_rewrite} and passage re-ranking~\cite{han2020learningtorank,passagererankingbert,nogueira2019multistage}. In the former, the current query and previous utterances are provided as input to generate the rewritten query, and in the latter, transformer-based models are fine-tuned on a relevance classification task to then score candidate passages.
%In the experiments, we show that this combination leads to state-of-the-art performance on TREC CAsT 2019.

\vspace{2mm}
\noindent\textbf{Knowledge-guided conversation response-generation.} The dialogue context can be captured by tracking the conversation knowledge over the different turns. Then, given the previous utterances' context and the current query, a natural language search-answer needs to be generated. In chit-chat dialogue generation agents, most approaches use encoder-decoder neural architectures that first encode utterances, and then the decoder generates a response~\cite{reinforcement_dialog_generation,adversarial_dialogue_generation,retrieval_based_generation_1,knowledge-aware-dialogue-generation,retrieval_based_generation_2}. 
For knowledge-guided generation, it is necessary to bias the generator such that it attends to knowledge-specific aspects, such as entities, that convey the conversation context. In a standard setting, models are trained end-to-end and the type of answers generated is entirely dependent on the training data. An interesting approach to bias answer generation, is retrieval-based dialogue generation, in which the generator takes as input retrieved candidate documents to improve the comprehensiveness of the generated answer~\cite{retrieval_based_generation_1,retrieval_based_generation_2}. In~\cite{query_biased_abstractive_summarization} a different approach is used to bias the generator, in which to obtain query-biased responses using a recurrent neural network, a copy mechanism is used to pay special attention to overlapping terms between the document and query.
In end-to-end dialog systems that incorporate external knowledge to generate answers, a common approach is to fuse information from a knowledge-base in encoder-decoder models~\cite{qin-etal-2019-entity,knowledge-aware-dialogue-generation}.
All these end-to-end approaches require a large dataset with annotated dialogues. In an open-domain conversational search setting, this is not feasible, as collections can be comprised by millions of passages. An alternative is to leverage on transfer-learning and use Transformer-based models pre-trained on large corpora, that have proved to be effective at abstractive summarization~\cite{t5}.
We depart from previous work by leveraging on pre-trained transformer models for a knowledge-guided answer-generation.  Given the knowledge encoded in entities relations over queries and passages, with demonstrated usefulness on search settings~\cite{entity_query_expansion,context_ranking_entities,entity_saliency_retrieval}, we propose to use the entities conversation graph to select a set of top passages that are fed to the generator, towards enforcing entity knowledge-graph bias in the generated answers.

\section{Conversation-aware Passage Retrieval}
\label{retrievalpassages}
%move this part to the introduction?
In \cite{castoverview} the conversational search task is defined as, given a sequence of natural language conversational query turns $T={q_1,...q_i,...q_n}$, the conversational search task aims to find the relevance passages that fit the current conversational context.

\begin{figure*}[t]
  \centering
    {\includegraphics[width=0.75\linewidth, trim=1pt 1pt 1pt 1pt, clip]{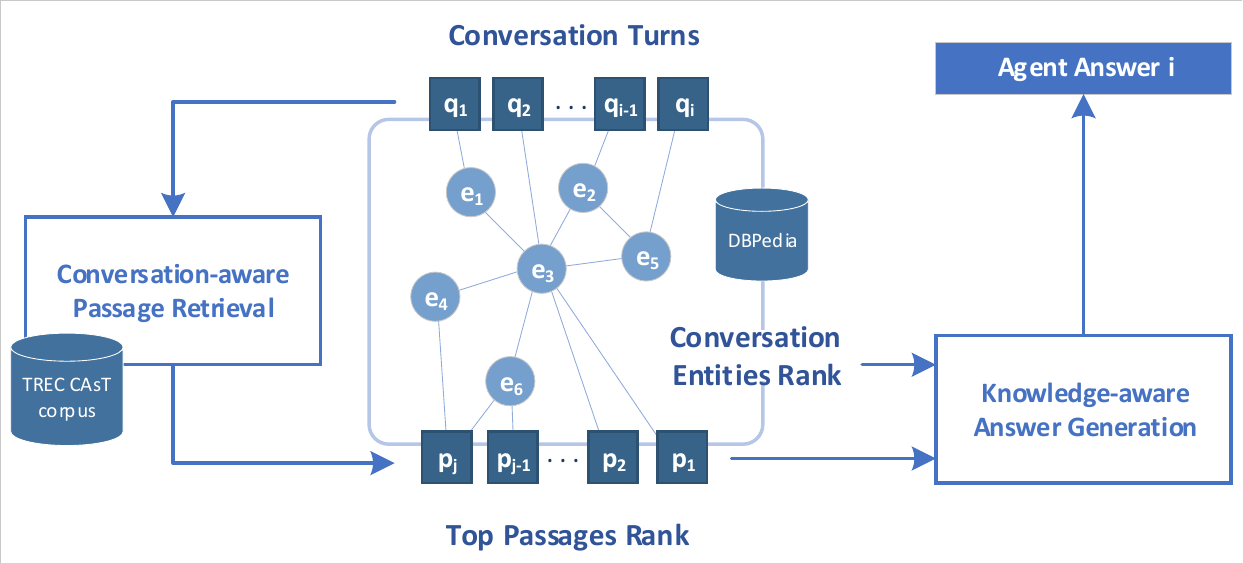}}
  \caption{Overview of the knowledge-driven conversational system and answer generation architecture.}
  \label{fig:conversationalarchitecture}
\end{figure*}

We implemented a three-stage conversation-aware passage retrieval pipeline composed of a context tracker, a first-stage retrieval and a re-ranker.
Because of the conversational characteristics of this task, the current query may not include all of the information needed to be answered.
To solve this, we use as the context tracking component, a query rewriting method based on the T5 model~\cite{t5} . This model requires an input sequence and a target sequence given as strings. Following \cite{t5conversational}, we fine-tune a T5-BASE model by providing as input the sequence of conversational queries and passages, and as target the rewritten query. In particular the input is defined as
\begin{equation}
\label{t5_input}
    ``q_i \ [CTX] \ q_1 \ p_1 \ [TURN] \ q_2 \ p_2 \ [TURN]  \ \ldots \ [TURN] \ q_{i-1} \ p_{i-1}",
\end{equation}
\noindent
where $i$ is the current turn, $q$ is a query, $p$ is a passage retrieved from the index by the retrieval model, and $[CTX]$ and $[TURN]$ are special tokens. $[CTX]$ is used to separate the current query from the context (previous queries and passages) and $[TURN]$ is used to separate the historical turns (query-answer pair).

The first-stage retrieval component uses a query likelihood retrieval model~\cite{languagemodelsmoothing} to recover a small set of passages. After the first-stage retrieval step, we re-rank the top-\textit{n} retrieved passages to obtain a better rank using a BERT model~\cite{bertOriginal}. This model generates contextual embeddings for a sentence and each of its tokens. We used a model fine-tuned on the passage ranking task~\cite{passagererankingbert} through a binary relevance classification task, where positive examples are relevant passages, and negative examples are non-relevant passages. 
To obtain the embeddings for a passage $p$ and a query $q$, BERT is fed with the following sequence of size $N$ tokens 
\begin{equation}
\label{eq:embedding_bert}
    emb = BERT(``[CLS]\ q \ [SEP] \ p"),
\end{equation}
\noindent
where $emb \in \mathbb{R}^{N \times H}$ ($H$ is BERT embedding's size), represents the embeddings of all tokens, and \textit{[CLS]} and \textit{[SEP]} are special tokens in BERT's vocabulary, representing the classification and separation tokens, respectively.
We then extract from $emb$ the embedding of the first token, which corresponds to the embedding of the \textit{[CLS]} token, $emb_{[CLS]} \in \mathbb{R}^{H}$.
This embedding is then used as input to a single layer feed-forward neural network (FFNN), followed by a \textit{softmax}, to obtain the probability of the passage being relevant to the query:
\begin{equation}
\label{eq:bert_passage_raking}
    P(p|q)=softmax( \text{FFNN}(emb_{[CLS]}) ).
\end{equation}
\noindent
With $P(p|q)$ calculated for each passage $p$ given a query $q$, the final rank is obtained by re-ranking according to the probability of being relevant.

\section{Knowledge-aware Answer Generation}
\label{sec_answer_gen}
In this section we address the key research hypothesis of this paper and propose a method to generate search-answers while considering the intersection between the entities in the top retrieved passages and the entities in the conversation turns.

The graph of entities is built from the top retrieved passages and queries from previous turns. 
Figure~\ref{fig:conversationalarchitecture} illustrates the rationale of the proposed approach.
We extract the entities from passages and queries, and then propose two methods to explore this information: one based on the relation of entities in the query and in the passage; and one that adapts the PageRank algorithm to a graph of entities.
The answer is then generated with the passages that exhibit a stronger relation with the most salient entities of the conversation until a given turn.

\subsection{Entity Linking}
To build the conversation-specific knowledge graph, we start by performing Entity Linking (EL) over both conversation queries and passages.
Entity linking tackled the two main existing classes of entities~\cite{entitydefinition}: named entities and concepts. 
The named entities class include specific locations, people and organisations. 
Concepts are abstract objects that include, but are not limited to, mathematical, physical and social concepts such as, ``distance'', ``gravity'' and ``authority''. 
We examined several entity linkers which are focused only on named entities (e.g. AIDA~\cite{ELaida} and FOX~\cite{fox}),  and on both named entities and concepts (e.g. WAT~\cite{ELwat} and DBpedia Spotlight~\cite{dbpediaspotlight}).

\subsection{Selection of the best Passages}
Each passage is scored according to the entity graph of the conversation. The relation between a query $q_i$, on turn $i$, and a candidate passage $p^k$, is computed as the average,
\begin{equation}
PassageScore(p^k| q_i) = \sum_{e_j \in E_{p^k}} \frac{EntityRank(e_j)}{\#|e_j|},
\label{eq:graph_score}
\end{equation}
where the target passage $p^k$ is scored as the sum of the $EntityRank(\cdot)$ scores of all entities $e_j$ present in that passage. When $E_{q}=\emptyset$, it is equivalent to $\gamma=0$ on equation~\ref{eq:graph_score}.
%we select the top-3 retrieved passages, otherwise, we select the top passages according to this equation~\ref{eq:graph_score}.

\subsection{Entities with Strong Pairwise-Relations}
Given that EL provides us with meaningful DBpedia identifiers for the mentions detected in text, we can obtain the relationship between two given entities, by exploiting their connections on DBpedia. This knowledge can be used to rearrange the order of the top passages provided by the previous step of our conversational system.

We can obtain a measure of \textit{entity relatedness} between $e_1$ and $e_2$, two entities of interest, following the measure proposed by Milne et al.~\cite{relatednessMeasure}:
\begin{equation}
EntRel(e_1,e_2|KB) = \frac{log(max(|E_1|,|E_2|) - log(|E_1\cap E_2|))}{log(|D|)-log(min(|E_1|,|E_2|))},
\label{eq:relatedness}
\end{equation}
where $E_1$ and $E_2$ are the sets of all entities that link to $e_1$ and $e_2$ in the KB, respectively, and $D$ is the set of all the entities in the KB. We use DBpedia~\cite{dbpedia} as our KB.

Using the $relatedness$ measure, for every turn $i$, we rearrange the top passages by considering the relatedness between the set of entities of top passages, $E_{p}$, and the set of query entities from the current and past turns, $E_{q}=\cup_{j=0}^{j=i} E_{qj}$, where $E_{qj}$ denotes the entities of query $q$ on turn $j$. By considering both the current and previous queries of the same conversation topic, we cover possible topic shifts in the conversation. The score of a passage according to the DBpedia entity relatedness, is computed as the average entity relatedness between a query $q_i$, on turn $i$, and each candidate passage $p^k$,
\begin{equation}
\begin{split}
PassageScore&(p^k|KB,q_i) =\\ 
&\frac{1}{|E_{q}||E_{p^k}|}\sum_{e_{1} \in  E_{q}}^{}\sum_{e_{2} \in E_{p^k}}^{} EntRel(e_{1},e_{2}|KB),
\end{split}
\label{eq:doc_score}
\end{equation}
where each current passage entity $e_1\in E_{p^k}$, is measured against every query-entity $e_2\in E_{q}$, from previous and current turns. When $E_{q}=\emptyset$, we select the top-3 passages.

\subsection{Conversation Knowledge Graph}
\label{subsec:graph_scoring}
In this section, we propose to look at the entire graph of entity relations that are in the top passages and query. This introduces a step change in relation to the above approach where we only looked at the relations between entities in the query and entities in each individual passage. 

To represent the conversation specific knowledge, we collect all the linked entities that occur in the queries and the top passages. This results in the concatenation of the query entities vector and the matrix with the entities in the passages,
\begin{equation}
\setlength{\dashlinegap}{2pt}
Map_E = 
\left[\begin{array}{c:c}
\gamma \cdot 
\begin{bmatrix}
q_{e_1} \\ 
\vdots\\
q_{e_n}
\end{bmatrix}
& (1-\gamma) \cdot 
\begin{bmatrix}
p^1_{e_1} & \multicolumn{2}{c}{\ldots} & p^m_{e_1}\\
 \vdots & \multicolumn{2}{c}{\ddots} & \vdots \\
p^1_{e_n} & \multicolumn{2}{c}{\ldots} & p^m_{e_n}
\end{bmatrix}
\end{array}
\right]
\end{equation}
where each element of the matrix is a boolean indicator of entity presence, and the $\gamma$ variable adjusts the importance of the entities in the queries vs passages. Moreover, to allow soft context shifts within the conversation, we consider the entities present in both the current and previous queries.

The graph of entities of a given conversation is computed as 
\begin{equation}
Graph_E = Map_E \cdot {Map_E}^T.
\end{equation}
This results in the covariance matrix between different entities.
We control the sparsity of the graph by cancelling entity relations below a given threshold. This allows us to compute the centrality of each entity $e_i$ in the conversation by applying the PageRank algorithm to the conversation entity-graph:
\begin{equation}
\begin{array}{ll}
EntityRank(e_i) &= \frac{1-\alpha}{N} 
\\&+ \alpha\cdot\sum_{e_j\in{neighbors(e_i, Graph_E)}} \frac{EntityRank(e_j)}{\#|e_j|},
\end{array}
\end{equation}
where $e_i$ is the target entity, $e_j$ correspond to a neighboring entity of $e_i$, $Graph_E$ is the conversation's entity graph, and the damping factor $\alpha$ was set to $0.99$.
The rationale for using PageRank, is that in the top passages there will be a stronger focus on the entities that are central to the conversation, while the entities that lie outside the conversation topic will be sparsely connected to the other entities in the graph.

\subsection{Answer Generation with Entity Relatedness}
Having identified a set of candidate passages according to the retrieval model (eq.~\ref{eq:bert_passage_raking}) and the entities knowledge, the goal is to generate a natural language response that combines the information comprised in each of the passages.
To address this problem, we follow an abstractive summarisation approach, which unlike extractive summarisation that just selects existing sentences, can portray both reading comprehension and writing abilities, thus allowing the generation of a concise and comprehensive digest of multiple input passages. Therefore, we select the passages that maximise the expression: 
\begin{equation}
\label{summarization_input}
 \argmax_{p^k} PassageScore(p^k,q_i) 
\end{equation}
\noindent
and generate the agent response with the sequence of the top $N=3$ passages, $``p^1\  p^2\  \ldots\  p^N"$. With this strategy, we implicitly bias the answer generation by asking the model to summarise the passages that are not only deemed as more relevant according to the retrieval system, but also that maximise the relatedness measure from eq.~\ref{eq:doc_score} or eq.~\ref{eq:graph_score}. 

This task has been commonly addressed by seq2seq models that learn to map input sequences to output sequences, but the Transformer architecture~\cite{vaswani2017attention} has led to groundbreaking results, due to its high effectiveness at modelling large dependency windows of textual sequences. Thus, in this work we consider the Text-to-Text Transfer Transformer (T5)~\cite{t5} based on the encoder-decoder Transformer architecture. This model is pre-trained on the large C4 corpus, which was derived from Common Crawl\footnote{\url{https://commoncrawl.org/}.}. A masked language modelling objective is used, where the model is trained to predict corrupted randomly sampled tokens of varying sizes.

\section{Evaluation}
\label{sec_evaluation}
\begin{figure*}[ht!]
    \centering
    \includegraphics[width=0.329\textwidth]{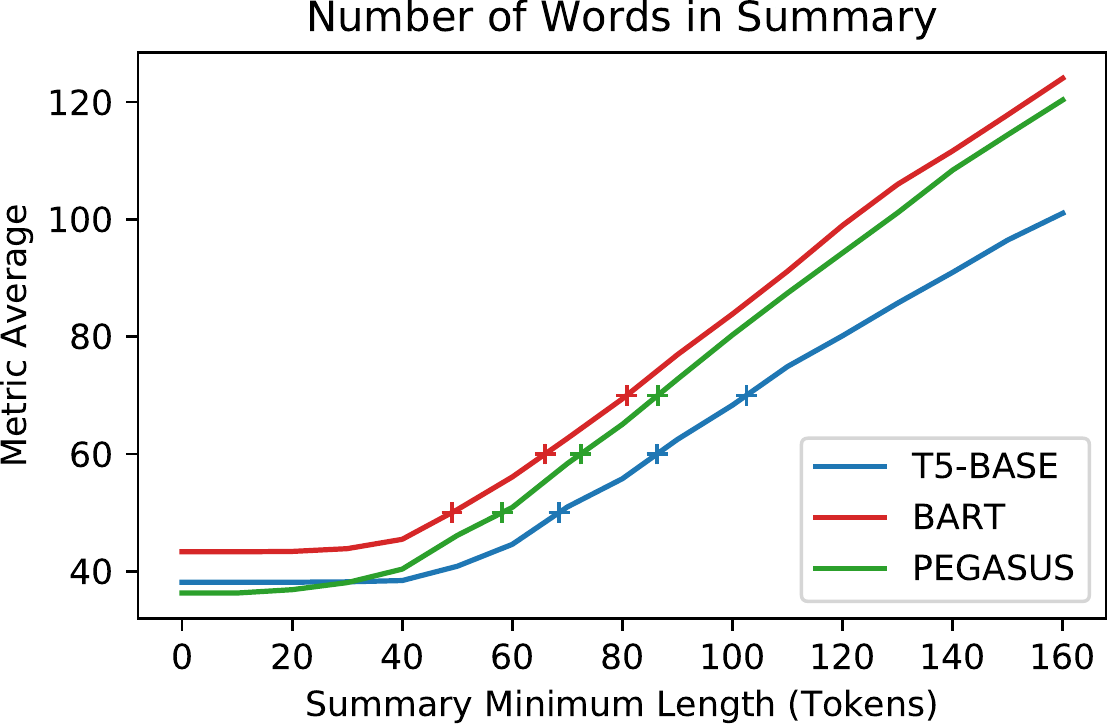}
    \includegraphics[width=0.329\textwidth]{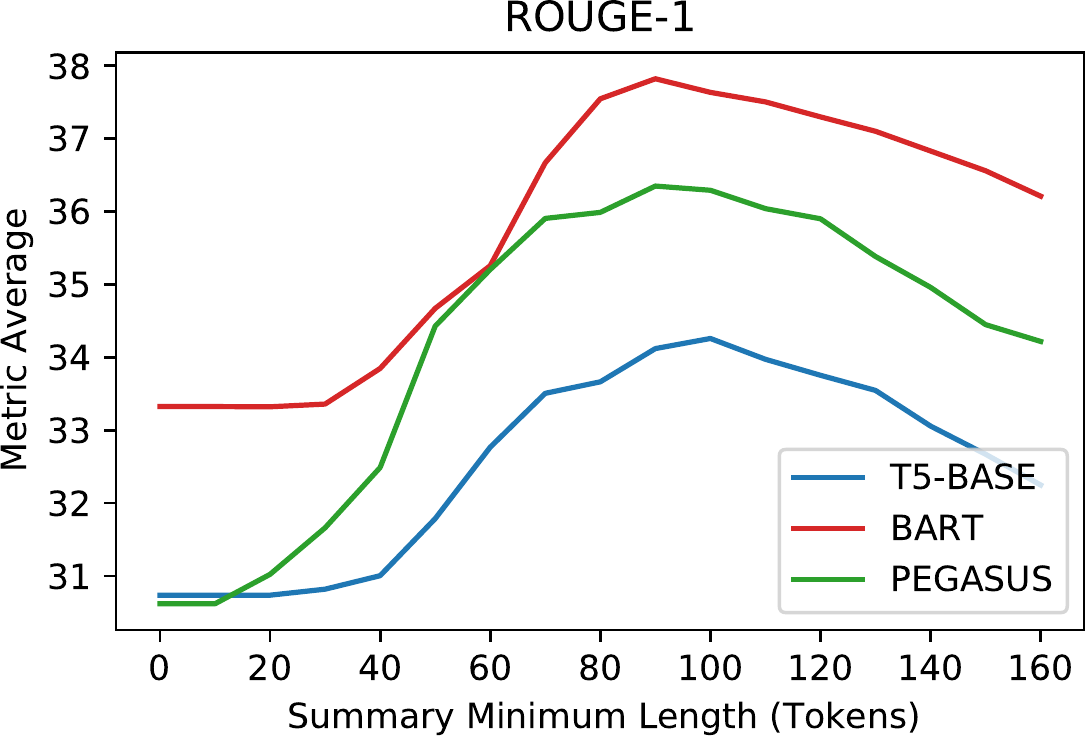}
    \includegraphics[width=0.329\textwidth]{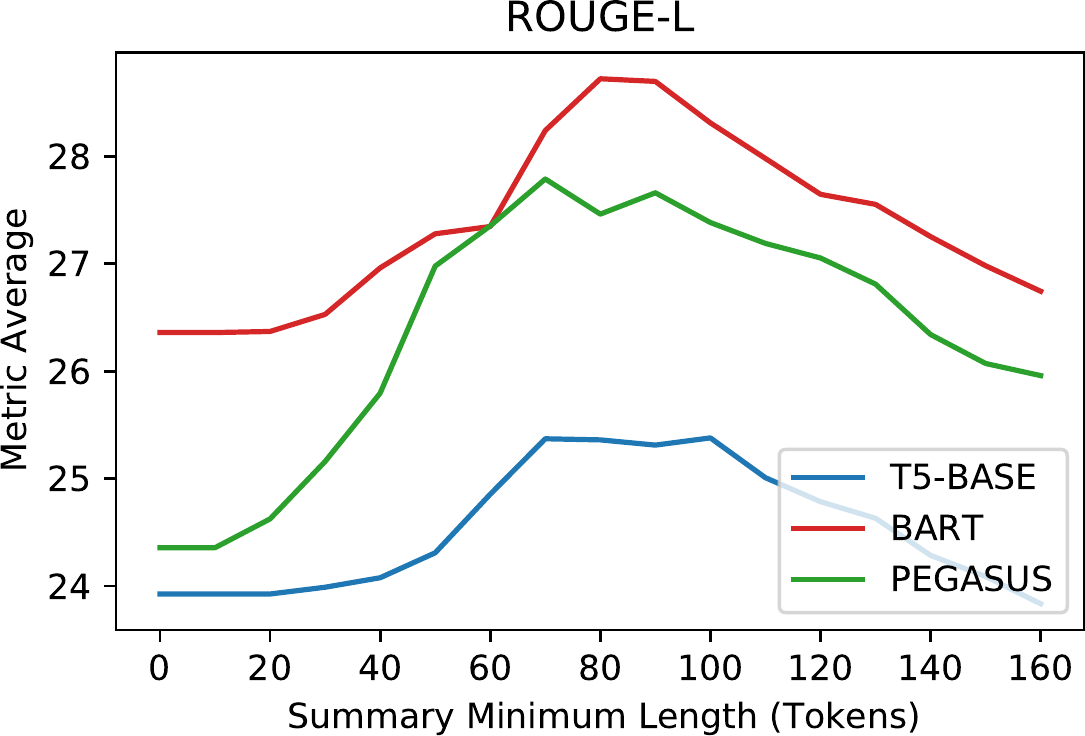}
     \includegraphics[width=0.329\textwidth]{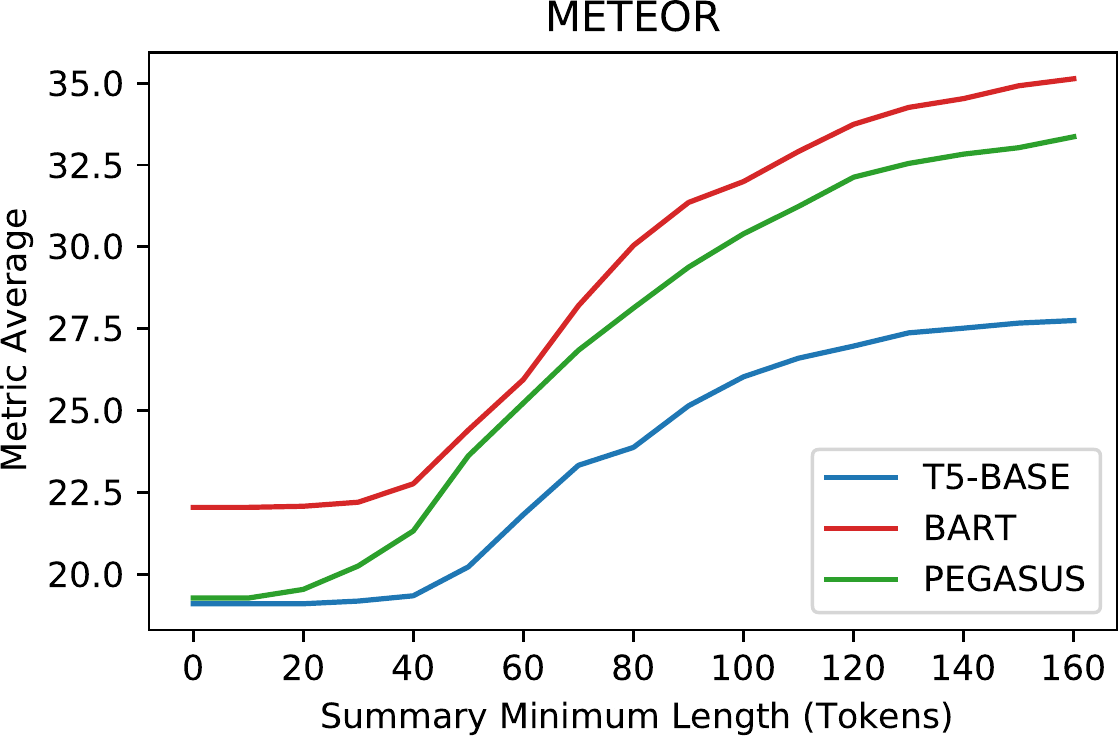}
    \includegraphics[width=0.329\textwidth]{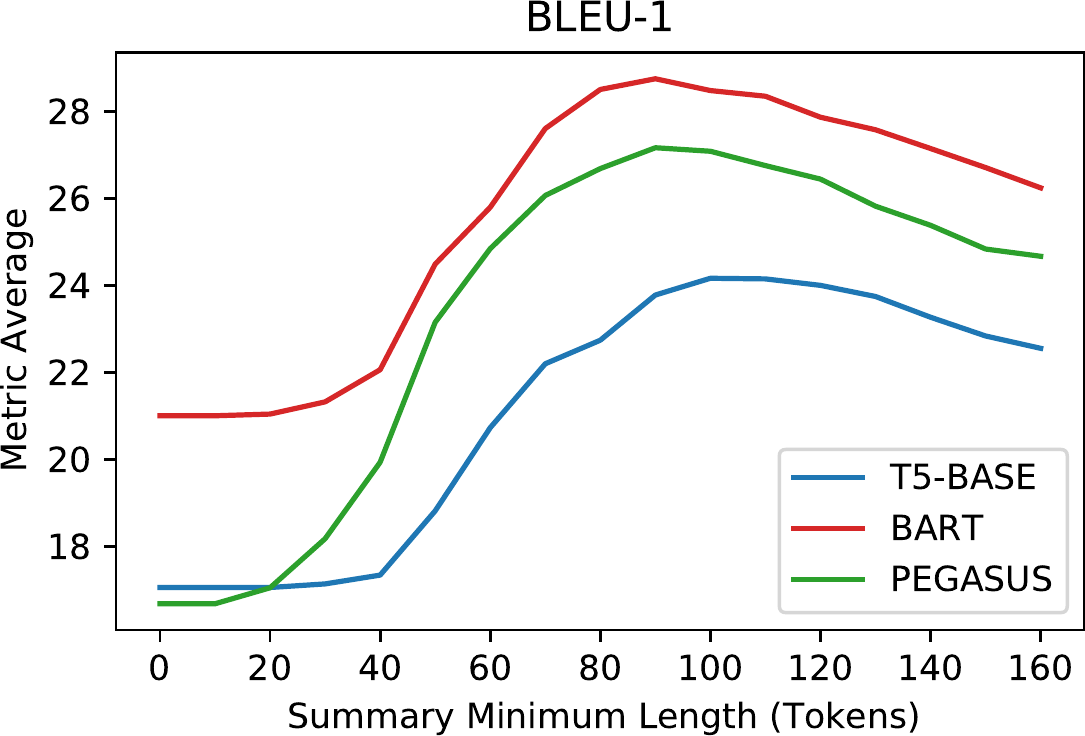}
    \includegraphics[width=0.329\textwidth]{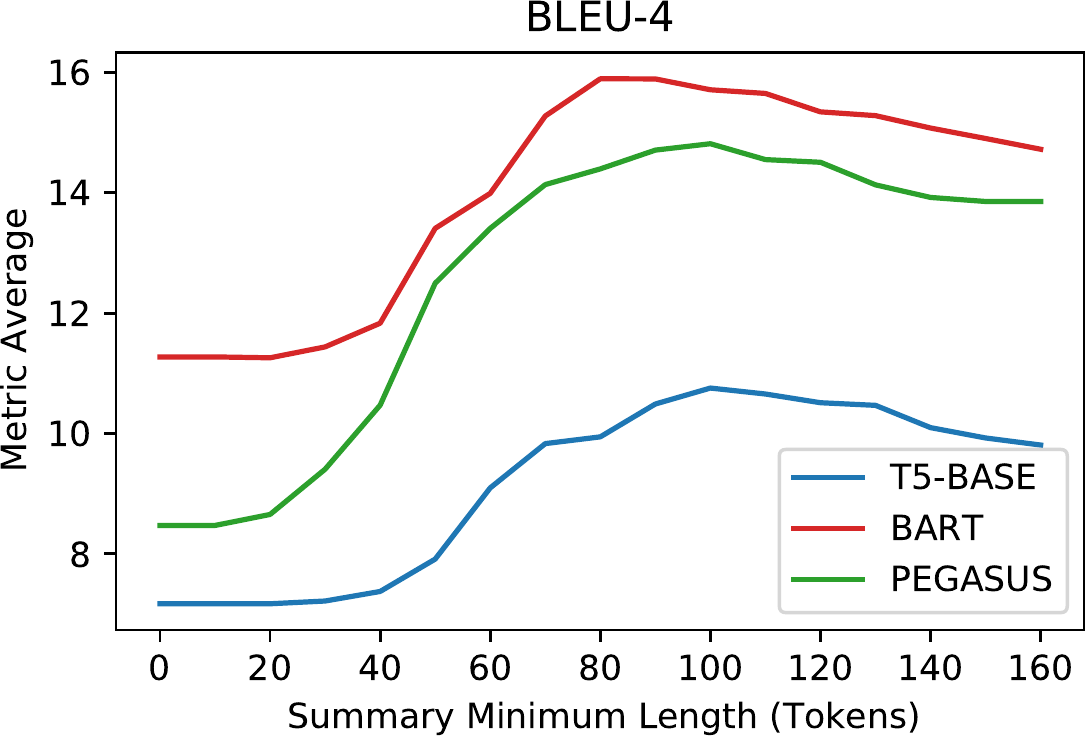}
    \caption{Showcase of how the \texttt{min\_length} parameter can influence the number of words of a summary and its metrics values.}
    \label{fig:nrwordsandmetrics}
\end{figure*}

\subsection{Datasets and Protocol}
%\subsubsection{CANARD Dataset~\cite{canYouUnpackIt}.}
%This dataset was used to train and evaluate the query rewriting method - It was created by manually rewriting the questions in QuAC~\cite{quac_dataset} to form non-conversational queries. The training, development, and test sets have 31.538, 3.418, and 5.571, query-rewrites respectively.

\subsubsection{TREC CAsT Dataset} The TREC CAsT dataset, ~\cite{trecCast}, was used to evaluate both the conversational retrieval and the knowledge-aware answer generation components. 
%To the best of our knowledge this is the only dedicated conversational search dataset.
There are 20 labelled conversational topics each with about 10 turns. The evaluation process uses a graded relevance that ranges from 0 (not relevant) to 4 (highly relevant).
The passage collection is composed by MS MARCO~\cite{marcoDataset}, TREC CAR~\cite{treccar}, and WaPo~\cite{washingtonDataset} datasets, which creates a complete pool of close to 47 million passages.

\subsubsection{Experimental Protocols}
% Computing BLEU: http://ssli.ee.washington.edu/~mhwang/pub/loan/bleu.pdf
%To analyse query rewriting performance, we used the BLEU-4 score~\cite{bleu} between the model’s output and the queries rewritten by humans, on the CANARD dataset.

To evaluate the passage retrieval component we used the TREC CAsT setup and the official metrics, nDCG@3 (normalised Discounted Cumulative Gain at 3), MAP (Mean Average Precision), and MRR (Mean Reciprocal Rank).
%In the passage retrieval experiment, we used the TREC CAsT setup and the official metrics, , along with Recall and P@3 (Precision at 3).

In the answer generation experiment, we used METEOR and the ROUGE variant ROUGE-L. 
The reference passages correspond to all the passages with a relevance judgement of 3 and 4. Hence, the goal is to generate answers that cover, as much as possible, the information contained in all relevant passages, in one concise and summarised answer.

\subsection{Implementation}

\subsubsection{Passage Retrieval}
\label{answer_retrieval_implementation}
To index and search we used Pyserini\footnote{\url{https://github.com/castorini/pyserini}.} and in specific the Language Model Dirichlet (LMD)~\cite{languagemodelsmoothing} retrieval model with the stemming algorithm Kstem\footnote{\url{http://lexicalresearch.com/kstem-doc.txt}.}. To perform re-ranking, we used a BERT LARGE model fine-tuned on a binary relevance classification task on the MS MARCO dataset~\cite{marcoDataset} following~\cite{passagererankingbert}. The query-rewriting component uses a T5-BASE model~\cite{t5} fine-tuned on the conversational query-rewriting task using the CANARD dataset~\cite{canYouUnpackIt}, following \cite{t5conversational}.

\subsubsection{Entity Linking}
For Entity Linking, we use DBpedia Spotlight~\cite{dbpediaspotlight} (DBS) to link general concepts and Named Entities, and the Federated Knowledge Extraction Framework~\cite{fox} (FOX) to link only Named Entities. 

\subsubsection{Transformer based answer generation}
To generate the answers that summarise the entity-focused passages, we employed the T5-BASE (T5) summariser~\cite{t5}\footnote{\url{https://huggingface.co/models}}, fine-tuned on the summarisation task with the CNN/Daily Mail dataset~\cite{cnndailymail}. To generate the summary, we use 4 beams, restrict the n-grams of size 3 to only occur once, and allow for beam search early stopping when at least 4 sentences are generated. We fix the maximum length of the summary to be of the same length of the input given to the models (3 passages) and vary the minimum length from 20 to 100 words.

\subsubsection{Parameters}
We observed that the value of the \texttt{min\_length} parameter is not directly proportional to the number of words in the created summary. We can see in Figure~\ref{fig:nrwordsandmetrics} an example showcasing this phenomena. For this example, the original top-3 passages are fed to the models. We select on the first graph the points corresponding to summaries of length 50, 60 and 70. We see that the different models, in order to create summaries of the same length between them, need different values for the \texttt{min\_length} parameter. Moreover, we can observe that summaries created with a similar number of words by the different models hold really different contents, as we can see in the graphs of the showcased metrics.

We can also see that the METEOR metric, much like the other applied metrics, follows the tendency established by the first graph. The ROUGE and BLEU metrics show to reach a peak in performance when setting the \texttt{min\_length} parameter to around 80, whose value is probably related to the average length of the reference summaries used in the evaluation procedure.

In order to properly compare the different annotators we then fix the number of words of our liking and extract the \texttt{min\_length} value that each different model requires to allow the creation of a summary with that attribute.

It is obvious that, in order to obtain the maximum possible value in METEOR, for instance, a big value for the \texttt{min\_length} parameter has to be chosen, and to obtain a better value in ROUGE and BLEU, the \texttt{min\_length} should be set to around 80 tokens, as easily seen in the Figure. However, by doing so, we argue that the goal of studying the generation of short and informative answers, essential for a conversational search setting, is completely missed. We believe that answers which possess the fewer words possible without losing information are the most desirable. With a quick analysis on the results yet to be presented we observed that 1) PEGASUS is the model which can present the shorter summaries out of the three models in all settings and that 2) the least number of words that all models can generate collectively applying all the proposed methods is on average 50. Because of this, we will fix the number of the summaries generated to have on average 50 words in order to better compare the different summarizers in this a setting where the least number of words possible is used to answer to a query.

\section{Results and Discussion}
\label{sec_results}
\subsection{Quantitative Results}

For all experiments we report the F1 scores for ROUGE-1, ROUGE-2 and ROUGE-L, because both precision and recall are important for the setting at hands. Precision shows to be important regarding the concise nature of the created summaries and Recall captures how much of the reference summary is captured in the created summaries. We find important to report ROUGE-2 scores in conjunction with ROUGE-1 to show the fluency of the created summaries, with the intuition that the more closely the words ordering of the reference summary is followed, the more fluent the summary can be considered.

\subsubsection{Answer Generation Baselines}
In order to assess the performance of the proposed models, we firstly measured the various summaries produced by varying the \texttt{min\_length} parameter from 0 to 160 tokens.

In order to study the performance of the summarizers, we established different baselines, targeting both the original rank of documents and the proposed methods to complement and aid a comparison between the results achieved by the different models. From this point forward, we will refer the proposed methods Entity Relatedness and Entity Graph Passage Scoring by their initials: ER and EG, respectively. We will refer from now on the ranked passages given by the model described in Section~\ref{retrievalpassages} as ``O'', which stands for Original passages rank.

The baselines present in Table~\ref{tab:baselinestop3} are composed by the top-3 text passages. We can clearly see that the baseline which shows better metric values has neither the biggest or smallest number of words. It is also relevant to point out that these baselines have at least the double of words than their Top-1 counterparts but not always present better metric values.

\begin{table*}[ht]
\caption{Averaged metric values for different baselines with top-3 passages.}
\centering
\begin{tabular}{@{}lcccccccc@{}}
\toprule
\textbf{Baseline} & 
\textbf{\# Words} &
\textbf{ROUGE-1} & 
\textbf{ROUGE-2} &
\textbf{ROUGE-L} &
\textbf{BLEU-1} & 
\textbf{BLEU-4} & 
\textbf{METEOR} 
\\ \midrule
Top-3 O & 237.21 &31.95 & 23.42 & 28.19 & 20.83 & 15.26 & 42.62\\
Top-3 ED & \textbf{205.50} & 28.88 & 17.28 & 23.54 & 18.62 & 10.97 & 34.82\\
Top-3 ER & 292.68  & 30.76 & 23.23 & 27.39 & 19.83 & 15.07 & 42.85 \\
Top-3 EG $\gamma=0$ & 319.75 &  30.24 & 23.52 & 27.34 & 19.44 & 15.31 & 43.14 \\
Top-3 EG $\gamma=.25$ & 282.72 &31.66 & 24.64 & 28.74 & 20.49 & 16.17 & 44.07\\
Top-3 EG $\gamma=.5$ & 287.73 & \textbf{32.23} & \textbf{25.47} & \textbf{29.37} & \textbf{20.88} &\textbf{ 16.69 }& \textbf{45.35}\\
Top-3 EG $\gamma=.75$ & 325.30 & 29.91 & 22.99 & 26.79 & 19.08 & 14.83 & 42.75\\
Top-3 EG $\gamma=1$ & 237.41 & 31.85 & 23.27 & 28.06 & 20.74 & 15.15 & 42.44\\
\bottomrule
\end{tabular}
\label{tab:baselinestop3}
\end{table*}

Before showing our main experiments results, we reaffirm that the usage of full text passages as references brings an unfair comparison between the created summaries and the above baselines. The baselines shown bring forward the evidence that having the biggest number of words does not led to better metrics and, ignoring the different number of input words, the baselines with best performance are achieved by following the proposed EG method. A more complete evaluation would use ground truth, however, this is the only feasible way regarding the dataset at hands. To compensate for this we also elaborated a human evaluation experiment whose results will be explored  further ahead .

\subsubsection{Answer Generation with Original Rank}

We will now report the results of our first experiment, following the original ranked passages, feeding the top-3 to the different models and fixing the \texttt{min\_length} parameter as needed to create summaries with 50 and 70 words on average. 

\begin{figure}[ht]
    \centering
    \includegraphics[width=0.5\textwidth]{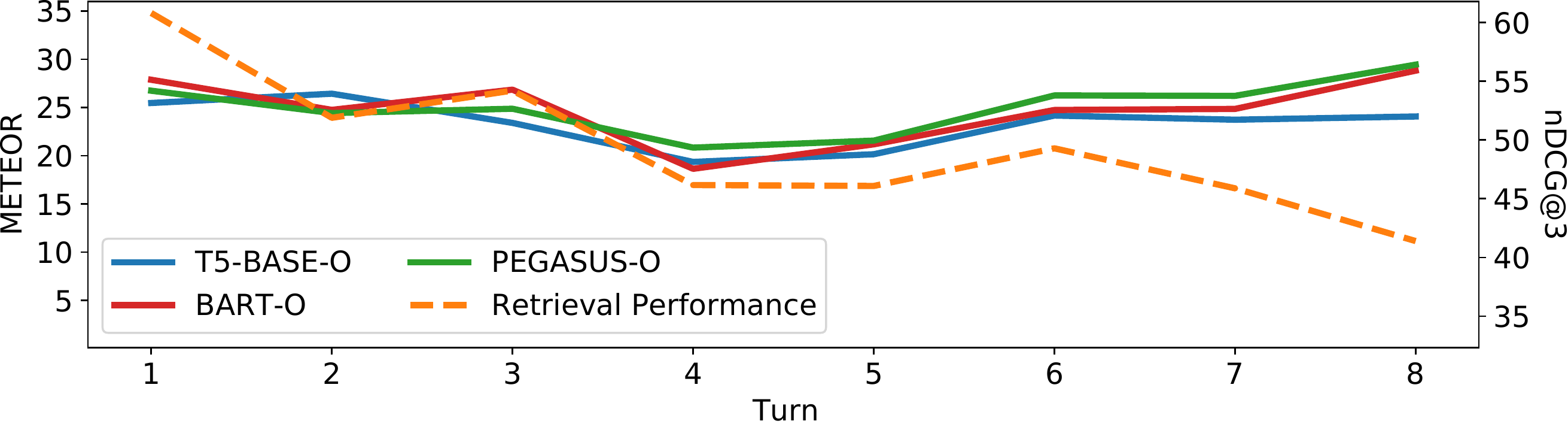}
    \includegraphics[width=0.5\textwidth]{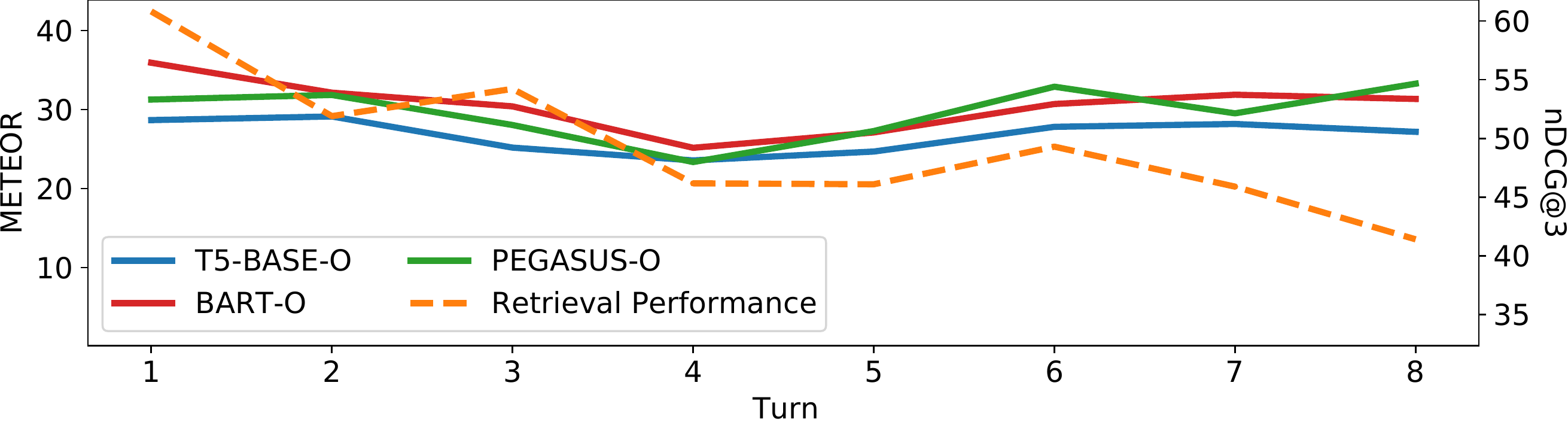}
    \caption{Answer generation versus retrieval performance per conversation turn. The average summary size is 50 and 70 in the top and bottom graphs respectively.}
    \label{fig:answerGenOverTurns}
\end{figure}

% In the bottom half of Table~\ref{tab:none50} we can see that all models, with the exception of BART, benefited from the addition of the query to the top-3 documents to be fed to the models. With this step it was needed to address the cases when the whole query was included in the models answer by filtering it out, since it is completely unnatural to receive an answer which includes the submitted query amidst it. Interestingly, T5 never included the query in the generated answer, while BART and PEGASUS showed to include the query in 4 distinct turns. BART showed to be more prone to include small queries while PEGASUS included lengthier ones.

In Figure~\ref{fig:answerGenOverTurns} we analyze the retrieval and the answer generation performance over conversation turns by making usage of two axis showcasing different metrics. We see that peak retrieval performance is achieved on the first turn, which was expected given that the first turn is the one that establishes the topic. As the conversation progresses, retrieval performance decreases, but surprisingly, answer generation performance is overall stable. When asked to perform summaries with 70 words as average, the METEOR values displayed become less dense. We also observed that the decreases in performance are linked to sub-topic shifts within the same conversation topic and BART is the model which tends to follow more closely the trend established by the retrieval performance. We suspect this is because of his extractive behavior.

Finally, in Table~\ref{table:answer_generation_example} we illustrate the answer generation with all the three Transformers. This Table further confirms the abstractive versus extractive summarization behaviors of the different Transformers. In this example we see that T5-BASE tries to generate new sentences by combining different sentences and PEGASUS makes use of verb synonyms not seen in text in order to convey the same message but with fewer words.

%For knowledge-guided answer generation we considered four baselines: (i) the top-3 most relevant passages obtained directly from the context-aware re-ranker of eq.~\ref{eq:bert_passage_raking} (BASELINE), (ii) the summary obtained from T5 using the same previous top-3 passages (T5-BASELINE), (iii) knowledge-driven summary using T5 and the top-3 passages according to the entity relatedness score from eq.~\ref{eq:doc_score} (T5-ER), and (iv) the knowledge-driven summary using T5 using the top-3 passages according to the entity graph score from eq.~\ref{eq:graph_score} (T5-EG). In all experiments, we report the results for different summary lengths.

\begin{table*}[t]
\centering
\caption{Answer generation example for the turn \textit{"What was the first artificial satellite?"}. 
The summaries have on average 50 words. Green sentences illustrate abstractive and blue sentences illustrate extractive summaries.}
\label{table:answer_generation_example}
\begin{tabular}{p{0.1\textwidth}p{0.85\textwidth}}
\toprule
% \textbf{Query Turn} &  What was the first artificial satellite? \\ \hline
\textbf{Method} &  \textbf{Answer} \\ \hline
\textbf{Retrieval Passage 1} &  The first artificial Earth satellite was Sputnik 1. Put into orbit by the Soviet Union on October 4, 1957, it was equipped with an on-board radio-transmitter that worked on two frequencies: 20.005 and 40.002 MHz. Sputnik 1 was launched as a step in the exploration of space and rocket development. While incredibly important it was not placed in orbit for the purpose of sending data from one point on earth to another. And it was the first artificial satellite in the steps leading to today's satellite communications. \\
\textbf{Retrieval Passage 2} &  The first artificial satellite was Sputnik 1. It was the size of a basketball and was made by the USSR (Union of Soviet Socialist Republics) or Russia. It was launched on October 4, 1957. \\ 
\textbf{Retrieval Passage 3} &  The first artificial satellite was Sputnik 1, launched by the Soviet Union on October 4, 1957, and initiating the Soviet Sputnik program, with Sergei Korolev as chief designer (there is a crater on the lunar far side which bears his name). This in turn triggered the Space Race between the Soviet Union and the United States. \\ \hline
\textbf{T5-BASE} & \textcolor{ForestGreen}{the first artificial satellite was launched by the ussr or Russia. it was the size of a basketball and launched on October 4, 1957.}  \textcolor{blue}{it was equipped with an on-board radio-transmitter that worked on two frequencies.}\textcolor{blue}{ it was not placed in orbit for the purpose of sending data from one point on earth to another.} \\ \hline
\textbf{BART} & \textcolor{blue}{The first artificial satellite was Sputnik 1, launched by the Soviet Union on October 4, 1957. It was equipped with an on-board radio-transmitter that worked on two frequencies: 20.005 and 40.002 MHz. This in turn triggered the Space Race between the Soviet Union and the United States.} \\ \hline
\textbf{PEGASUS} & \textcolor{ForestGreen}{Sputnik 1 was launched by the Soviet Union on October 4, 1957.} \textcolor{blue}{It was the first artificial satellite in the steps leading to today's satellite communications.} \textcolor{ForestGreen}{It was not used to send data from one point on earth to another. Sputnik 1 triggered the Space Race between the Soviet Union and the United States.} \\
\bottomrule
\end{tabular}
\end{table*}

\subsubsection{Answer Generation with Entity Graph Rank}

\begin{figure*}[ht]

  \centering
  \includegraphics[width=.32\textwidth]{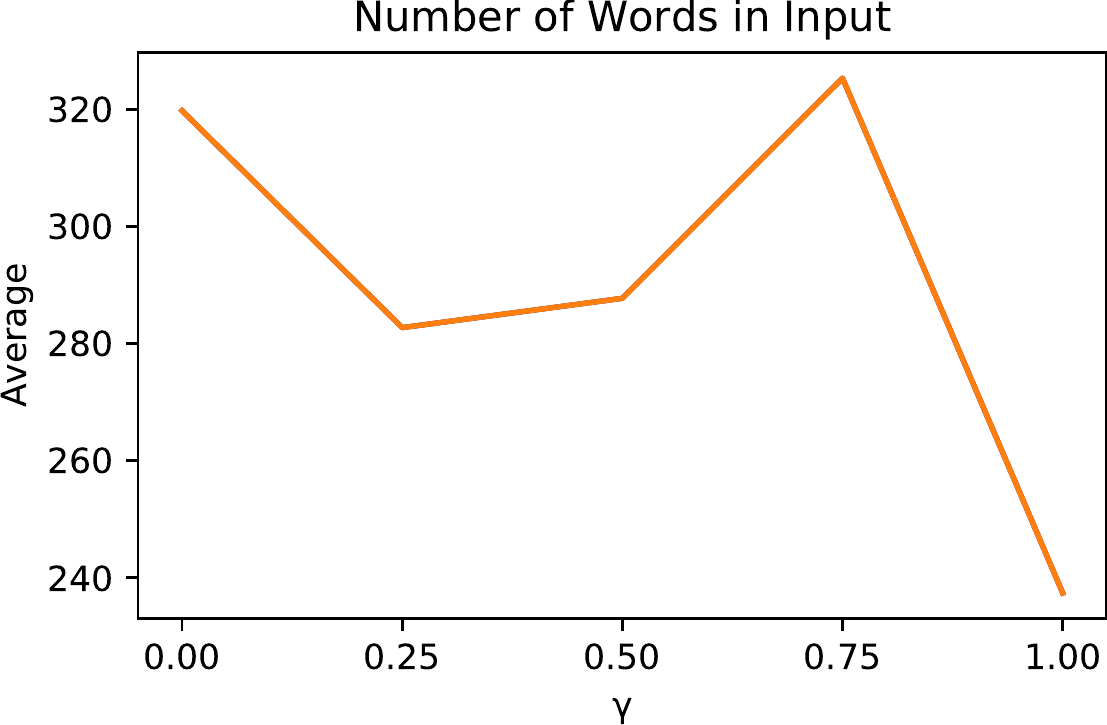}
\hspace{0.1cm}
  \includegraphics[width=.32\textwidth]{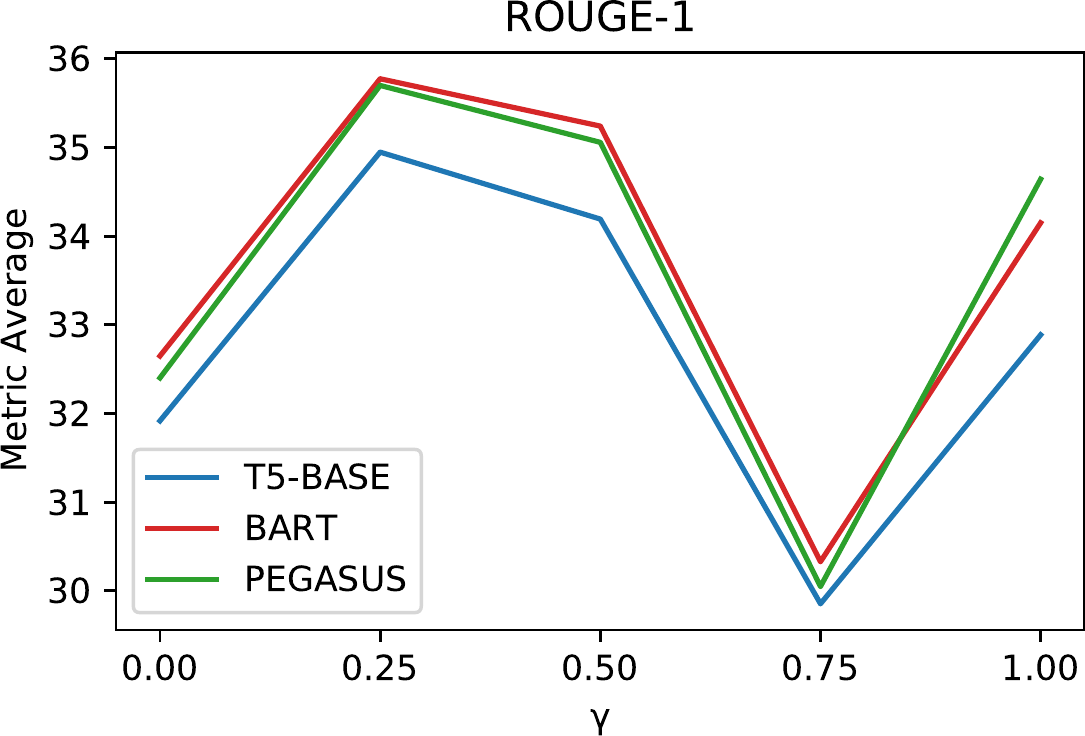}
\hspace{0.1cm}
  \includegraphics[width=.32\textwidth]{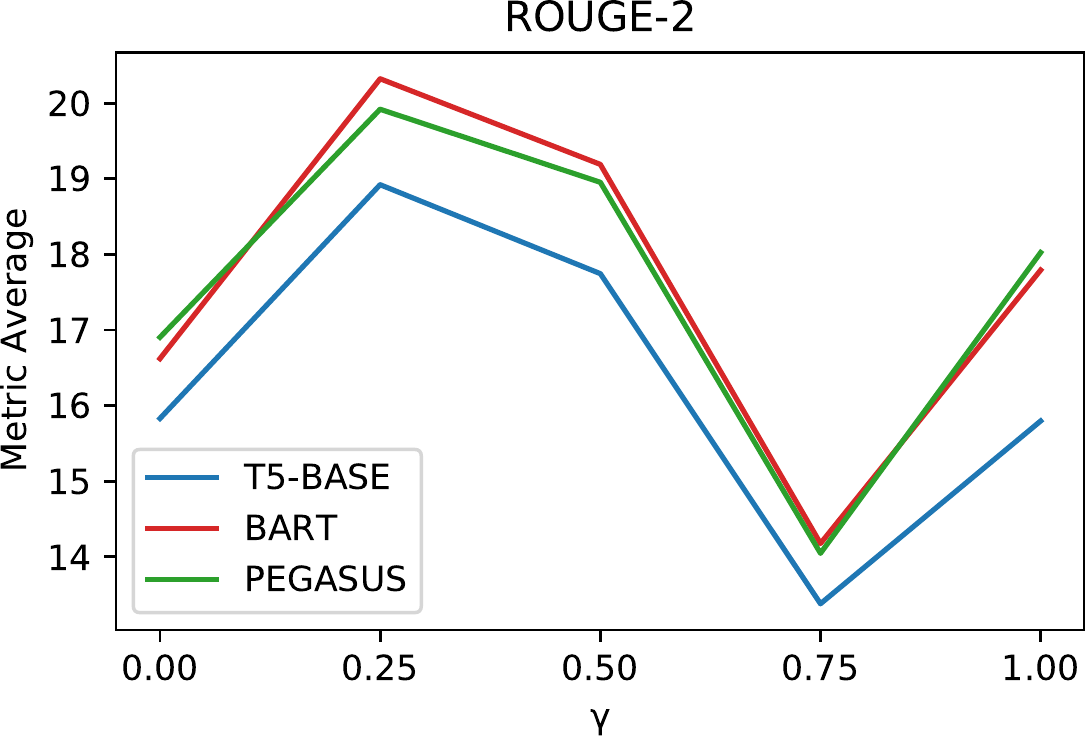}

\includegraphics[width=.32\textwidth]{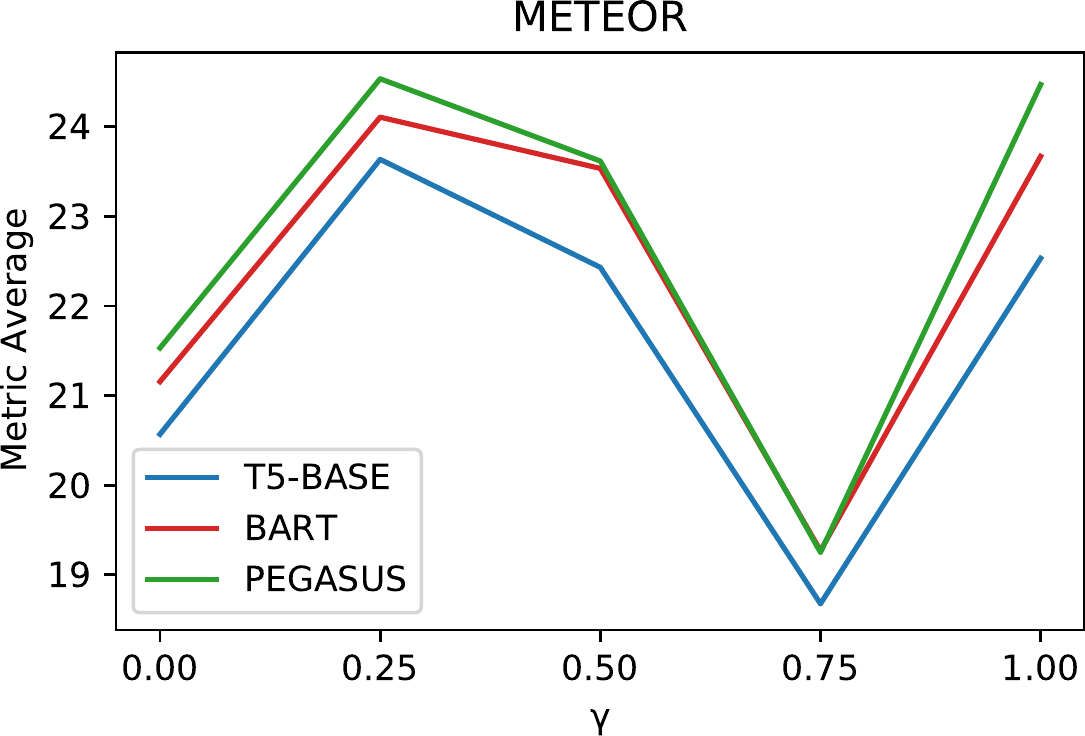}
\hspace{0.1cm}
  \includegraphics[width=.32\textwidth]{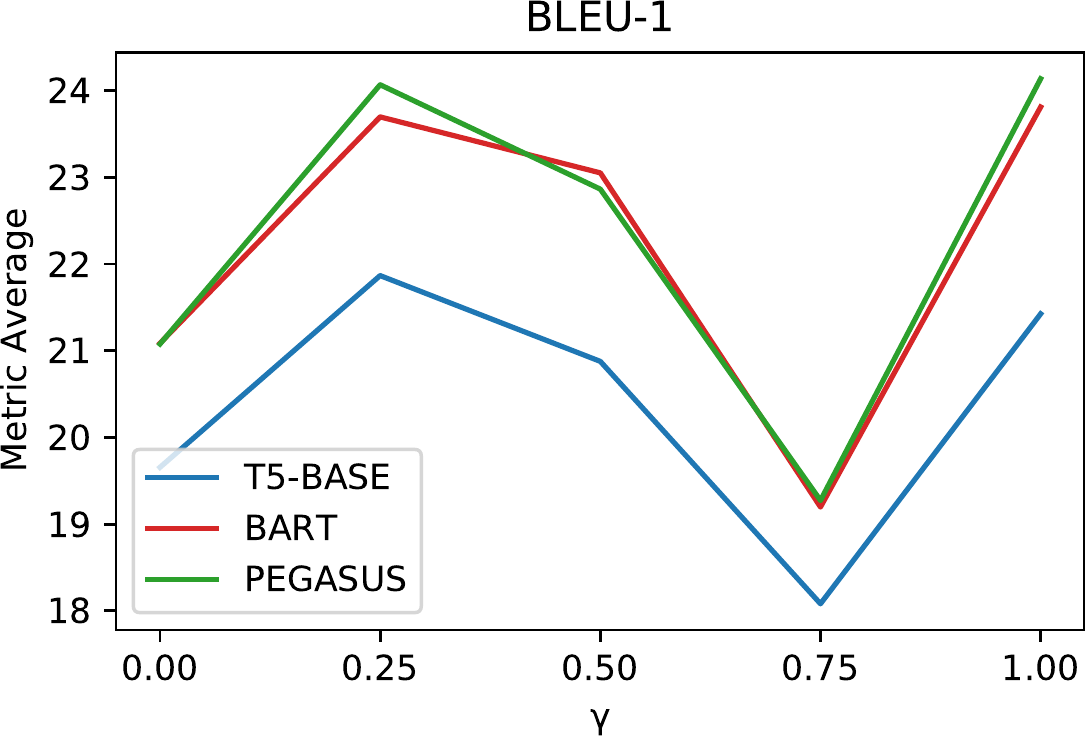}
\hspace{0.1cm}
  \includegraphics[width=.32\textwidth]{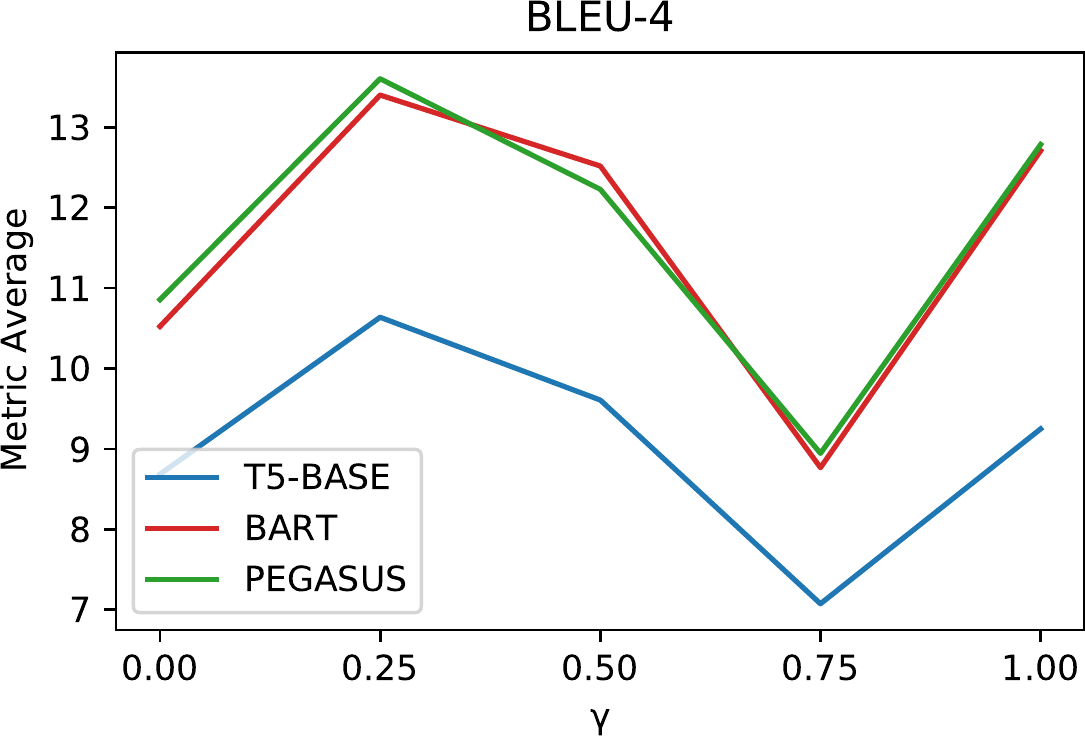}

  \caption{Input size and answer generation performance results under different $\gamma$ values for summaries of length 50.}
  \label{fig:answerGenGamma}
\end{figure*}

In the third experiment, we focus on the proposed Entity Graph Passage Scoring (Section~\ref{subsec:graph_scoring}), i.e. EG, and investigate the impact of $\gamma$ on the summary quality. Figure~\ref{fig:answerGenGamma} shows the summary generation quality results. On a quick glance it may appear as the scores are directly related to the number of words in the input, but with a more attentive inspection we can see that it is not quite right, as $\gamma=0$ and $\gamma=0.75$ have approximately the same number of words but induce different metrics performance. We observe that the best results are obtained with $\gamma=0.25$, meaning that more weight was given to entities from top-10 retrieved passages, but entities from current and previous queries still received some weight.

This supports our initial intuition that \textit{all} entities contribute to the context of the conversation. 
While the conversation-aware query rewriting solves coreferences on the current query, it is still important to consider entity relations between passages, and current and all previous queries. This is further evidenced by the results with $\gamma=1.0$ (passage entities are ignored), where it was the second best. When $\gamma=0.0$ (query entities are ignored) performance drops, as only knowledge from passage entities is used.

For entity graph passage scoring, we selected $\gamma=0.25$ based on the previous experiment. We observe that BART-EG overall achieves the best results although PEGASUS-EG shows to have better metric scores regarding BLEU and METEOR with summaries of 50 words. Also we notice that with this approach the addition of the query to the top-3 documents does not lead to a better performance.

\begin{table*}[ht]
\caption{Averaged metric values for summaries created with the top-3 documents following the Entity Graph method as input (top) plus query (bot).}
\centering
\begin{tabular}{@{}lccccccc@{}}
\toprule
\textbf{Model} & 
\textbf{\# W} &
\textbf{ROUGE-1} & 
\textbf{ROUGE-2} &
\textbf{ROUGE-L} &
\textbf{BLEU-1} & 
\textbf{BLEU-4} & 
\textbf{METEOR} 
\\
\midrule
%   Top-1 O Trim & 49.36 & 35.50 & 20.02 & 29.27 & 26.65 & 16.55 & 26.36\\
%  Top-1 O & 72.71 & 38.38 & 22.88 & 31.73 & 30.78 & 20.01 & 31.48 \\
 T5-BASE-ER  & & 31.97 & 15.07 & 24.43 & 19.97 & 8.23 & 21.13\\
 BART-ER & 50 & 33.83 & 17.75 & 26.53 & 22.95 & 12.22 & 22.93\\
 PEGASUS-ER  & & 32.15 & 15.86 & 25.14 & 21.18 & 10.41 & 21.54\\
\hdashline
T5-BASE-ER  & & 33.35 & 16.09 & 24.89 & 23.50 & 10.13 & 24.64\\
BART-ER  & 70 & \textbf{36.57} & \textbf{19.95} &\textbf{ 27.97} & \textbf{27.39} & \textbf{14.98} & \textbf{28.11}\\
PEGASUS-ER &  & 35.09 & 18.43 & 26.76 & 25.93 & 13.63 & 26.76\\
 \midrule
T5-BASE-EG & &  35.90 & 20.08 & 28.50 & 22.19 & 11.34 & 24.37\\
BART-EG  & 50 & 36.84 & 21.58 & 29.66 & 24.16 & 14.19 & 24.99\\
 PEGASUS-EG & & 36.89 & 21.36 & 29.18 & 24.63 & 14.64 & 25.52\\
\hdashline
 T5-BASE-EG & & 37.08 & 20.47 & 28.19 & 25.77 & 12.76 & 27.88\\
BART-EG & 70 &  \textbf{39.86} &\textbf{ 23.80 }& \textbf{31.11} & \textbf{30.17} & \textbf{18.46} & 30.85\\
 PEGASUS-EG &  & 39.50 & 23.36 & 30.76 & 29.52 & 17.90 & \textbf{30.89}\\
% \midrule
%  T5-BASE-EG-wQ &  & 35.09 & 18.59 & 27.04 & 22.11 & 10.21 & 23.65\\
% BART-EG-wQ & 50 & 36.06 & 20.83 & 28.86 & 23.73 & 13.86 & 24.48\\
%  PEGASUS-EG-wQ &  & 36.39 & 20.73 & 28.60 & 24.55 & 14.07 & 25.37\\
% \hdashline
% T5-BASE-EG-wQ &  & 36.19 & 19.03 & 26.76 & 25.28 & 11.68 & 26.94\\
%  BART-EG-wQ & 70 & 38.66 & 22.40 & 29.89 & 28.88 & 17.10 & 29.53\\
%  PEGASUS-EG-wQ &  & 38.16 & 21.83 & 29.78 & 28.35 & 16.43 & 29.44\\
%  \midrule

\bottomrule
\end{tabular}
\label{tab:eg}
\end{table*}

Looking at the Top-1 O Trim baseline results (appended at the end of Table~\ref{tab:eg}), we can easily notice that, for summaries with 50 words, the EG approach leads to better ROUGE-1 and ROUGE-2 scores from all models. BART and PEGASUS also show to have better ROUGE-L scores than this baseline. Regarding the original Top-1 O baseline and looking at the results achieved with $\#W=70$ we can also notice that PEGASUS and BART far surpass this baseline regarding ROUGE-1 and ROUGE-2. As for the other metrics, these models present really close results.

\begin{figure}[ht]
    \centering
  
    \includegraphics[width=0.5\textwidth]{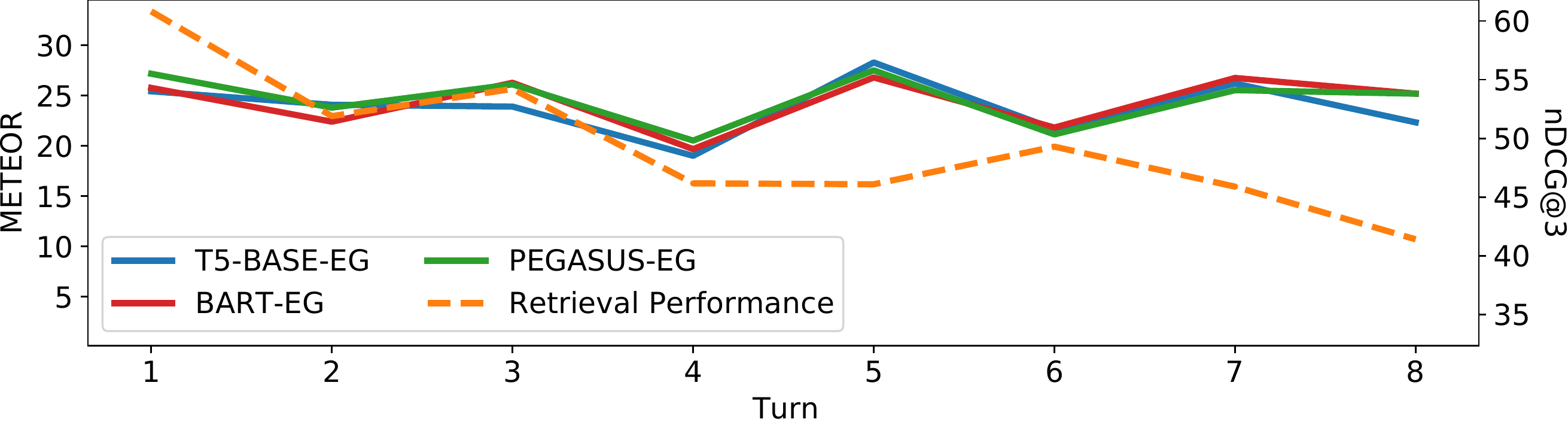}
    \includegraphics[width=0.5\textwidth]{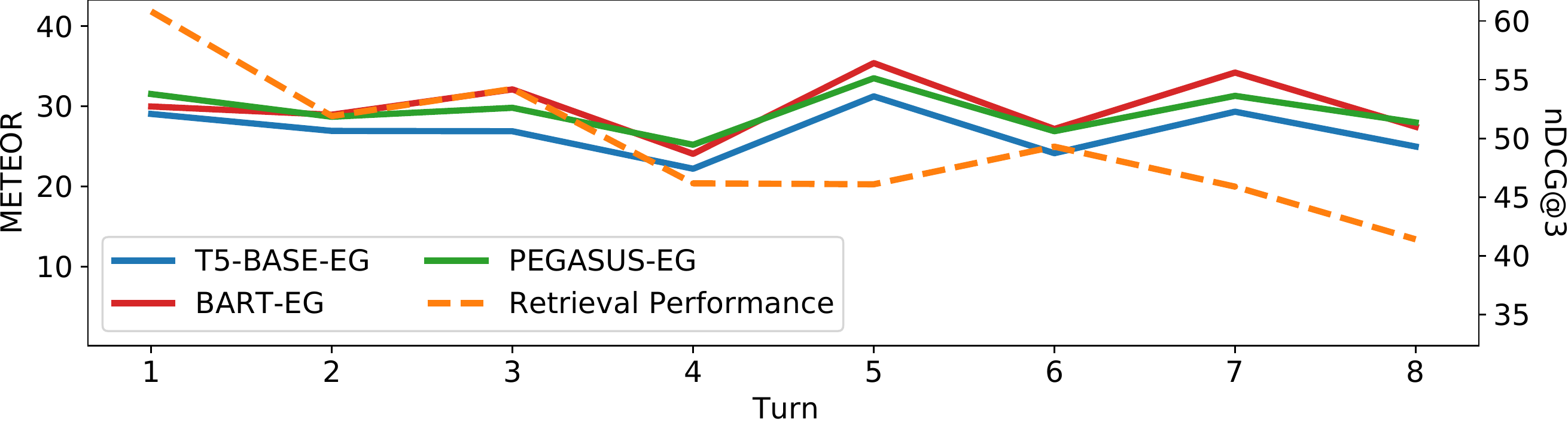}
    \caption{Answer generation versus retrieval performance per conversation turn applying the EG  method. The minimum length is 50 and 70 in the top and bottom graphs respectively.}
    \label{fig:answerGenOverTurnsEG}
\end{figure}

To show how the performance evolves through the different turns of the conversation, we can see in Figure~\ref{fig:answerGenOverTurnsEG} that the ``ER'' approach maintains a stable performance (although it is noticeable that the ``peaks'' achieved by ``ER'' bring an overall better performance). As it was observed with the ``O'' method, the different models performance is evidenced by creating summaries with more words. Interestingly, we can see that ``ER'' follows more closely the trend established by the retrieval performance, specially with summaries with size 70.

\subsection{Human Evaluation}

To better assess how the different proposed methods and baseline impact the information quality, conciseness and naturalness of the answers given in the conversations, we conducted a human evaluation experiment on Amazon Mechanical Turk. In this experiment we asked that each Worker would evaluate a conversation by rating each conversation turn on two 1-5 Likert scales, with higher being better, each targeting, for each turn:

\begin{itemize}
    \item Information Quality (IQ) - which aims to evaluate how well a answer addressed the query of the present turn, taking into account the context of the conversation.
    \item Naturalness and Conciseness (NC) - which aims to evaluate if the answers can be though as being created by human beings and don't include too much extraneous information.
\end{itemize}

% In all human evaluation experiments we used the same task template shown in Figure~\ref{fig:templateamt}. 
Each task comprised one random conversation created with a combination of model (baseline, T5-BASE, BART, PEGASUS), method (O, ER, EG) and length (50, 70). For the baseline the method is fixed as the O one and the length is not controlled. The user, of course, didn't know which combination was being asked to be evaluated in a given task. We chose to study only these combinations since encompassing all of them would result in the rise of the experiments' complexity.

% \begin{figure}[ht]
%     \centering
%     \includegraphics[width=\textwidth]{Figures/templateAMT.png}
%     \caption{A screenshot of the Amazon MTurk HIT which shows first and last conversation turn.}
%     \label{fig:templateamt}
% \end{figure}

Each task was independently done by 4 different Workers which, to be able to partake in the task, had to present a minimum approval rate of 95\% and had to at least have completed 100 Human Intelligence Tasks (HITs) already. Additionally, all HITs were inspected to the best of our ability and when, for a single user, a continuous session of HITs submission took place, the first and last submissions time and number of HITs performed was taken to calculate the average time spent per HIT. Users that showed an average value of less than 20 seconds had their submissions rejected and those HITs were re-submitted by other users. On total, this experiment was performed by 136 people. 
%We can see in Figure~\ref{fig:workers} that some Workers performed far more HITs than others. This Figure is not complete, not containing all the users which performed only one HIT, for ease of display. 

In a perfect situation, we argue that the number of HITs per Worker should have been more or less the same for all Workers, to take into account extreme Workers that consistently evaluate highly/low all HITs received. However, this feature is not accessible via the Amazon Mechanical Turk platform and we did verify that the results didn't drastically change when the top performing users were not pictured in the experiment results.

\begin{figure*}[ht]
    \centering
    \includegraphics[width=0.8\textwidth]{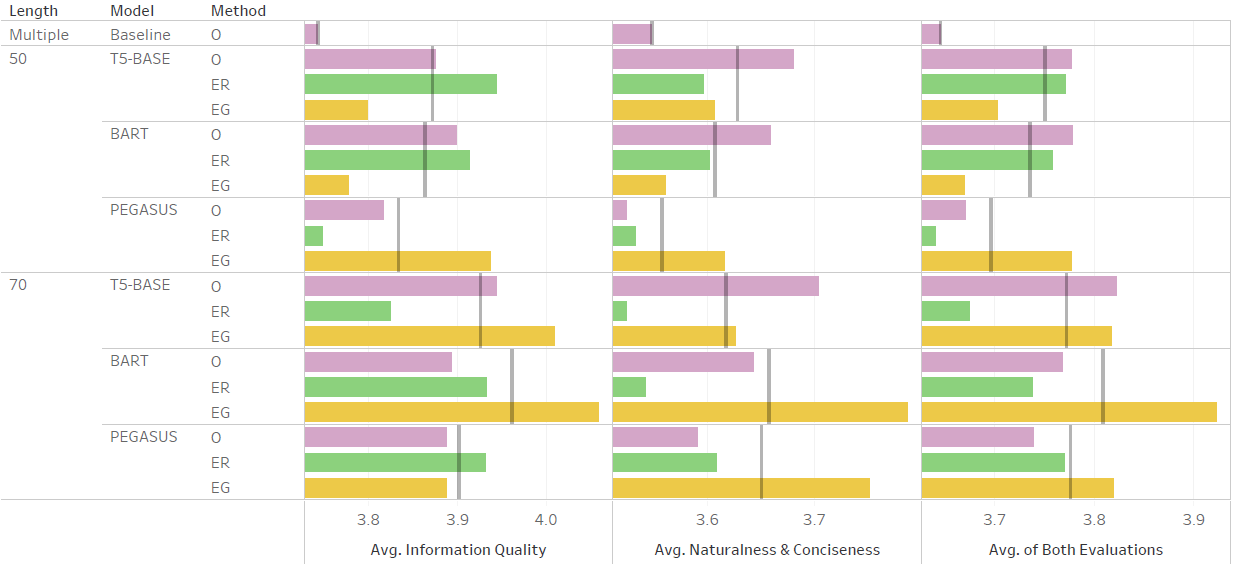}
    \caption{Averaged values of the human evaluation in terms of Information Quality, Naturalness and Conciseness per combination of Length, Model and Method. Each table pane shows a vertical gray line which represents the average of each Model with the different Lengths set.  }
    \label{fig:humaneval}
\end{figure*}

In Figure~\ref{fig:humaneval} we can see the obtained evaluation of Information Quality and Naturalness and Conciseness that were averaged per row. Each row comprises 20 different conversations that were evaluated each by 4 different Workers, totaling in 80 evaluations per row.

First off, we can observe that overall the Information Quality of the answers were rated higher than the Naturalness and Conciseness, as we can see in the range used in both axis. We must also be aware that the range in each axis is focused on portraying the differences between each score. If the axis would start at 0, the differences wouldn't be so striking, but nonetheless would still be there.

We can easily see that the Baseline shows the worst value in terms of IQ and the best result, both in terms of IQ and NC is presented with the combination of the BART model with answers of 70 words generated with the EG method. The EG method seems to synergize well with T5-BASE with the combination of also Length 70 in terms of IQ. In terms of NC, the results suggest that the method EG coupled with PEGASUS and with Length 70 also achieves better results than other combinations.

On the other hand, the summaries of 50 words created using PEGASUS and the ER method show, on both evaluation types, low results, suggesting that this combination should not be used to this conversational search setting.

% In Figure~\ref{fig:model_method_iq_nc} we show the mean values of IQ and NC when the results are aggregated by Model and Method. Easily we can again see the low score of the Baseline contrasting the higher values of the results shown by the Transformer Models. In terms of methods, we can see that the EG stays as the prime choice both in IQ and NC settings, while ER and O achieve second best results in IQ and NC settings, respectively.

% \begin{figure*}[ht]

% \centering
% \includegraphics[width=\textwidth]{Figures/model_IQ_NC.png}

% \includegraphics[width=\textwidth]{Figures/method_IQ_NC.png}
% \caption{Aggregated results by Model (top) and Method (bottom).}
% \label{fig:model_method_iq_nc}
% \end{figure*}

In order to better understand the final conclusions we can gather from this experiment and check if the results obtained reveal statistical significance, we performed ANOVA and paired t-tests with $\alpha = 0.05$.

With all the information gathered, these experiment results suggest the following:

\begin{itemize}
    \item \textbf{Summaries created with 70 words on average have better Information Quality.}
\end{itemize}

In order to determine if there is an interaction effect between our three independent variables (Model, Length and Method) on our continuous dependent variables (Information Quality and Naturalness and Conciseness) we performed a three-way ANOVA test. The results targeting both dependent variables only show statistical significance when the dependent variable studied was the Information Quality. 
%These results can be consulted in the Table~\ref{tab:3anova} and 
We observe that that indeed summaries created with more words (70) were preferable in comparison with summaries created with fewer words (50). 

This goes against our initial hypothesis that a Conversational Search answer should be short and informative but nonetheless this can make sense if our initial setting of 50 words is too short to convey the information that is required to met the conversation goal. This can also be connected with the type of questions that are asked, since there are questions that can be more straightforward answered than others. An other explanation can be that users of these systems want to know the most possible about the discussed topic, without it being overwhelming, of course.

This test also suggests that there was a statistically significant three-way interaction between Model, Length and Method, $F(4, 342) = 2.437, p = .047$.

\begin{itemize}
    \item \textbf{The best combinations to improve our Conversation Search system Information Quality wise are:}
    \begin{itemize}
        \item EG method with summaries of length 70 created by BART.
        \item EG method with summaries of length 70 created by T5-BASE.
        \item ER method with summaries of length 70 created by PEGASUS.
    \end{itemize}
\end{itemize}

We targeted the comparison of each combination of Model/Length/Method with the Baseline in order to better understand if the combinations indeed lead to a better system performance and were not achieved by mere chance. A paired-samples t-test was conducted to compare the IQ and NC (averaged per topic) separately, in the conditions where the combination of Model/Length/Method were and were not applied. 
%These results can be consulted in the Table~\ref{tab:humaneval05}.

Focusing the dependent variable IQ, there was a significant difference in the scores for the BART/70/EG $(M=4.05, SD=.33)$ and the Baseline $(M=3.74, SD=.31)$ conditions, $t(19)=-3.25, p = 0.004$. There was also a significant difference in the scores for the T5-BASE/70/EG $(M=4, SD=.32)$ and the Baseline conditions, $t(19)=-2.68, p = 0.015$. Additionally, a significant difference in the scores for the PEGASUS/70/ER $(M=3.93, SD=.3)$ and the Baseline conditions was found, $t(19)=-2.22, p = 0.039$. 

These results suggest that these combinations really do have an impact on the IQ perceived in the conversations. Specifically, our results suggest that when these combinations are applied, the systems performance in terms of Informational Quality increases.

In order to better understand if any of these combinations were indeed superior to the other two, we performed a second paired-samples t-test, but didn't find statistical relevant results both in terms of IQ and NC.

% \usepackage{booktabs}
\begin{table*}[ht]
\centering
\resizebox{\columnwidth}{!}{%
\begin{tabular}{@{}cclll@{}}
\toprule
      IQ &        & \multicolumn{1}{c}{}        & \multicolumn{1}{c}{Model} & \multicolumn{1}{c}{}        \\ \cmidrule(l){3-5} 
Length & Method & \multicolumn{1}{c}{T5-BASE} & \multicolumn{1}{c}{BART}  & \multicolumn{1}{c}{PEGASUS} \\ \midrule
       & O      & 77.5 (+2.6)                & 78.0 (+3.1)              & 76.3 (+1.5)                 \\
50     & ER     & 78.9 (+4.0)                & 78.3 (+3.4)              & 75.0 (+0.1)                \\
        & EG     & 76.0 (+1.1)                & 75.5 (+0.7)              & \underline{78.7 (+3.9)}                \\
        \midrule
       
        & O      & 78.9 (+4.0)                & 77.9 (+3.0)              & 77.8 (+2.9)                 \\
70     & ER     & 76.5 (+1.6)                & 78.7 (+3.8)              & \textbf{78.7 (+3.8)}       \\
       & EG     & \textbf{\underline{80.2 (+5.3)}}       & \textbf{\underline{81.2 (+6.3)}}     & 77.8 (+2.9)                  \\ \bottomrule
\end{tabular}
}
\resizebox{\columnwidth}{!}{%
\begin{tabular}{@{}cclll@{}}
\toprule
       NC        &        & \multicolumn{1}{c}{}        & \multicolumn{1}{c}{Model} & \multicolumn{1}{c}{}        \\ \cmidrule(l){3-5} 
Length & Method & \multicolumn{1}{c}{T5-BASE} & \multicolumn{1}{c}{BART}  & \multicolumn{1}{c}{PEGASUS} \\ \midrule
      & O      & 73.6 (+2.6)                & 73.2 (+2.2)              & 70.5 (+0.5)                \\
50     & ER     & 72.0 (+1.0)                & 72.1 (+1.1)               & 70.7 (+0.3)                \\
       
        & EG     & 72.15 (+1.2)                & 71.2 (+0.2)              & 72.3 (+1.4)                \\\midrule
      
       & O      & \underline{74.1 (+3.1)}                & 72.9 (+1.9)              & 71.81 (+0.8)                \\ 
70     & ER     & 70.5 (-0.5)                & 70.85 (-0.1)              & 72.2 (+1.2)                \\
        & EG     & 72.6 (+1.6)                & \underline{75.7 (+4.8)}              & \underline{75.0 (+4.9)}                \\ \bottomrule
\end{tabular}
}
\caption{Human evaluation side-by-side results on a 1-100 scale. Baseline reports the mean value of 74.85 in terms of IQ and 70.97 in terms of NC. Bold values are statistically significant, difference between baseline is shown between parenthesis in percentage and underlined values show the best score achieved by each model.
}
\label{tab:humaneval0100}
\end{table*}

It's interesting to notice that the best performing combinations make usage of the proposed methods. We can see in the Table~\ref{tab:humaneval0100} the results of the human evaluation in the scale of 0-100 in order to better visualize the differences between each combination. Regarding the BART/70/EG combination, we can see an improvement of over 6\%.

\section{Discussion}
\label{sec_discussion}

We deem useful to use automatic metrics (ROUGE, BLEU, METEOR) as proxies for measuring quantitatively the results achieved by the different models with the different proposed methods. However, these only provide limited information and don't give forward information regarding fluency and information needs met.

To this regard, options which were shown to not improve metric scores can still be considered for a further analysis since they can contribute for more natural and informative answers.

\subsection{Analysis of the Conversational Answers created with Added Query}

When using the query jointly with the top-3 passages, interesting cases arose in which the query was ``woven'' into the summary, producing a much natural and desirable answer. We can see an example of this phenomena in Table~\ref{tab:wQexample}.

\begin{table*}[ht]
\caption{Different answers given by the different model approaches for the question ``What is the largest shark ever to have lived on Earth?'' }
\centering
\begin{tabular}{p{0.45\textwidth} p{0.45\textwidth}}
\toprule
\textbf{PEGASUS-O} &
\textbf{PEGASUS-O-wQ} 
\\
\midrule
The megalodon is an extinct species of shark that roamed the waters of Earth over 1.5 million years ago. Although now extinct, it is still listed in the Guinness World Records as the largest shark (...) &  The largest shark to have ever lived on Earth is thought to have been the megalodon. Although now extinct, it is still listed in the Guinness World Records as the largest shark (...)\\
\bottomrule
\end{tabular}
\label{tab:wQexample}
\end{table*}

There are cases in which the addition of the query 1) does not lead to any different in the creation of the summary, 2) makes the summary use the same words but with a different ordering (usually starting with the words seen in the query) and 3) changes completely the summary created, with no similarity between the original and added query approaches.

\subsection{Analysis of the Conversational Answers with the Entity Density filter}
As with other methods, the application of the Entity Density filter does not automatically imply a different answer generation. Looking at the overall filtered passages, it is noticeable cases in which the filter acts as expected, as demonstrated in Table~\ref{tab:EDexample}.

\begin{table*}[ht]
\caption{Different processed text passages to give answer to a question about blood. }
\centering
\begin{tabular}{p{0.45\textwidth} p{0.45\textwidth}}
\toprule
\textbf{BART-EG} &
\textbf{BART-EG-ED} 
\\
\midrule
\underline{Confidence votes 133.} Red blood cells are produced in the bone marrow. Red blood cells are also known as erythrocytes (...) &  Red blood cells are produced in the bone marrow. Red blood cells are also known as erythrocytes (...)\\
\bottomrule
\end{tabular}
\label{tab:EDexample}
\end{table*}

However we observed cases in which the filter removed phrases which could bring relevant information forward. It was also noticeable that some top-3 texts gather a lot of irrelevant information forward but the ED method could not be applied, since the confidence parameter of Entity Linkers set could led to the identification of relevant concepts amidst the text. We believe that for texts with a bigger number of words this confidence parameter should be set higher in order to better curate information by being more critical about the contents to be selected.

We invite the reader to visit our Conversation Interactive Explorer\footnote{\url{https://knowledge-answer-generation.herokuapp.com}} and check the differences in quality that each available parameter can have in the showcased conversations.

\subsection{Analysis of a Conversation Knowledge Graph}
\begin{figure*}[ht]
    \centering
    \includegraphics[width=0.7\linewidth, trim=1pt 1pt 1pt 1pt, clip]{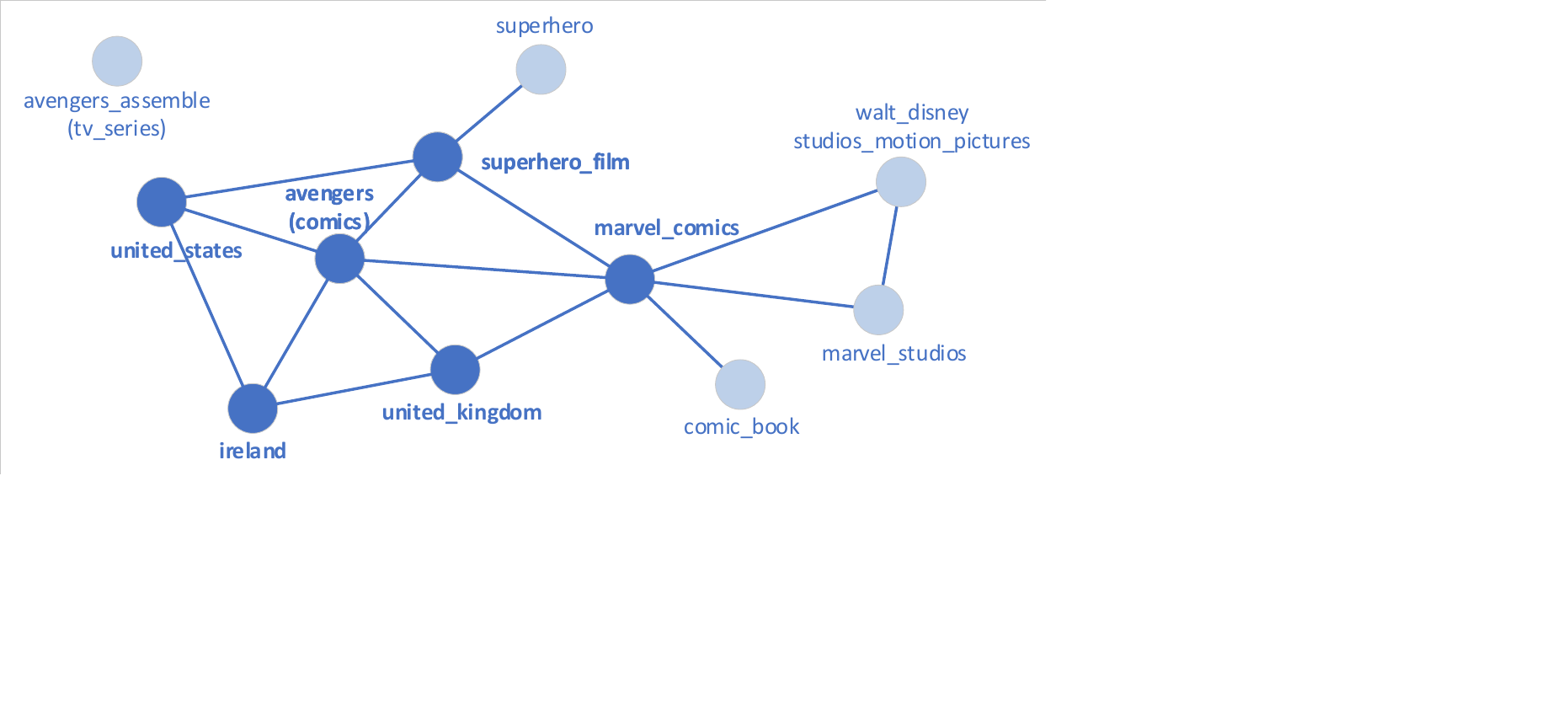}
    \caption{Conversation entity graph for the topic ``The Avengers''. The graph considers the most salient entities of the top-10 passages.}
    \label{fig:graph}
\end{figure*}

Figure~\ref{fig:graph} depicts the entity graph of the top-10 passages, from a conversation turn focused on the entity ``The Avengers''. The top-3 passages and corresponding answer summary made by T5-BASE can be seen in the Table~\ref{tab:top3andsummary}. From the Graph~\ref{fig:graph}, we can identify top entities (dark blue), the most salient entities for the current turn of the present topic, and bottom entities (light blue), which are connected to top entities but are not so central. As such, passages that have a better coverage of those entities (according to eq.~\ref{eq:graph_score}), are expected to be ranked higher. We can see that the 3 passages in Table~\ref{tab:top3andsummary} gather both top and bottom entities. Using these passages as targets to the answer generation component, we can see that the produced answer ends up accounting for all this information and successfully answers the given query. 

%These passages are more desirable to be ranked first since the probability of these entities being relevant is higher and 
%With this proposed re-ordering, the generated summary can successfully answer the requested question with the sought off attributes.
%We can clearly see that entities that most often co-occur are tightly connected to each other.
%We can clearly see that the most common entities are tightly connected to each other. 
%revealing that there is at least one passage in the top 10 that gathers all these entities, hence covering the common topics of the top 10 passages. 

\begin{table*}[ht]
\centering
\caption{Answer generation example. The summary minimum length is set to 90. The top and bottom entities are highlighted in blue and light blue respectively.}
\label{table:answer_generation_exampleEG}
\begin{tabular}{p{0.1\textwidth}p{0.85\textwidth}}
\toprule
\textbf{Query Turn:} &  Who are The Avengers? \\ \hline
\textbf{Passage 1} &  \textcolor{blue}{The Avengers} (2012 film) \textcolor{blue}{Marvel's The Avengers} (classified under the name \textcolor{blue}{Marvel} \textcolor{LightBlue}{Avengers Assemble} in the \textcolor{blue}{United Kingdom} and \textcolor{blue}{Ireland}), or simply \textcolor{blue}{The Avengers}, is a 2012 \textcolor{blue}{American superhero film} based on the \textcolor{blue}{Marvel Comics} \textcolor{LightBlue}{superhero} team of the same name, produced by \textcolor{LightBlue}{Marvel Studios} and distributed by \textcolor{LightBlue}{Walt Disney Studios Motion Pictures}. \\ 
\textbf{Passage 2} & \textcolor{blue}{ Marvel's The Avengers} (\textcolor{blue}{Marvel} \textcolor{LightBlue}{Avengers Assemble} in the \textcolor{blue}{UK} and \textcolor{blue}{Ireland}) more commonly known as \textcolor{blue}{The Avengers}, is a 2012 \textcolor{blue}{American} \textcolor{blue}{superhero film}, scripted and directed by Joss Whedon, based on the \textcolor{blue}{Marvel Comics} \textcolor{LightBlue}{superhero} team of the same name. The film stars an ensemble cast consisting of Robert Downey, Jr., Chris Evans, Mark Ruffalo, Chris Hemsworth, Scarlett Johansson, Jeremy Renner, Tom Hiddleston, Clark Gregg, Cobie Smulders, Stellan Skarsgård and Samuel L. Jackson.  In \textcolor{blue}{The Avengers}, Nick Fury (Jackson), director of the peacekeeping organization S.H.I.E.L.D., recruits Iron Man (Downey), Captain America (Evans), the Hulk (Ruffalo), and Thor (Hemsworth) to form a team that must stop Thor's adoptive brother Loki (Hiddleston) from subjugating Earth. \\
\textbf{Passage 3} & \textcolor{blue}{The Avengers} (also known as \textcolor{blue}{Marvel's The Avengers} and classified in the \textcolor{blue}{UK} and \textcolor{blue}{Ireland} under the title \textcolor{blue}{Marvel} \textcolor{LightBlue}{Avengers Assemble}) is a 2012 \textcolor{blue}{American superhero film} produced by \textcolor{LightBlue}{Marvel Studios} and distributed by \textcolor{LightBlue}{Walt Disney Studios Motion Pictures}, based on the \textcolor{blue}{Marvel Comics} \textcolor{LightBlue}{superhero} team of the same name. \\ \hline
\textbf{T5-BASE Answer} & \textcolor{blue}{The Avengers} is a 2012 \textcolor{blue}{american superhero film} based on the \textcolor{blue}{Marvel comics} \textcolor{LightBlue}{superhero} team of the same name. the film stars an ensemble cast consisting of Robert Downey, Jr., Chris Evans, Mark Ruffalo, Chris Hemsworth, Scarlett Johansson and Jeremy Renner. In the film, Nick Fury recruits Iron man, Captain America, the Hulk and Thor to form a team that must stop Loki from subjugating earth. \\
\bottomrule
\end{tabular}
\label{tab:top3andsummary}
\end{table*}

\section{Conclusions}
\label{sec_conclusions}
%chapter 3 conclusion remarks
In this paper we proposed a knowledge-aware answer generation method that considered the conversation specific graph of entities. The key findings of this paper are as follows:

\begin{enumerate}
\item \textbf{Knowledge-aware Search-Answer Generation.} The proposed method was able to abstract the information contained in multiple passages to generate a single, yet informative, search-answer.
This reduces the burden on the user that now only needs to read a snippet long answer containing links to multiple passages.

\item \textbf{Conversation-specific Knowledge-Graph.} The quality Knowledge-graph creation process is directly influenced by the quality of the retrieved passages. A fundamental step in the creation of Conversation Knowledge-graphs was a state-of-the-art conversational search baseline that we used to select the seeds for the graph creation process.

\item \textbf{Conversation-specific Rank of Entities.} The final critical element in the proposed method, concerns the ranking of entities by their importance during the conversation. We applied a modified PageRank algorithm to detect the salient entities along the conversation to focus the answer generation process in the corresponding passages.
b
\end{enumerate}

The results presented in this paper support the initial hypothesis and opened other questions that we plan to investigate in the future. The first one being related to the quality of entity-linkers, which may be further improved. The second concerns the research of models that can seamlessly combine the Transformer architecture advantages with the conversation knowledge-graphs.

\bibliographystyle{model2-names}
%https://github.com/tpavlic/splncs04nat
\bibliography{bibliography}

\begin{thebibliography}{42}
\expandafter\ifx\csname natexlab\endcsname\relax\def\natexlab#1{#1}\fi
\providecommand{\url}[1]{\texttt{#1}}
\providecommand{\href}[2]{#2}
\providecommand{\path}[1]{#1}
\providecommand{\DOIprefix}{doi:}
\providecommand{\ArXivprefix}{arXiv:}
\providecommand{\URLprefix}{URL: }
\providecommand{\Pubmedprefix}{pmid:}
\providecommand{\doi}[1]{\href{http://dx.doi.org/#1}{\path{#1}}}
\providecommand{\Pubmed}[1]{\href{pmid:#1}{\path{#1}}}
\providecommand{\bibinfo}[2]{#2}
\ifx\xfnm\relax \def\xfnm[#1]{\unskip,\space#1}\fi
%Type = Book
\bibitem[{Balog(2018)}]{entitydefinition}
\bibinfo{author}{Balog, K.}, \bibinfo{year}{2018}.
\newblock \bibinfo{title}{Entity-Oriented Search}. volume~\bibinfo{volume}{39}
  of \textit{\bibinfo{series}{The Information Retrieval Series}}.
\newblock \bibinfo{publisher}{Springer}.
\newblock \URLprefix \url{https://eos-book.org},
  \DOIprefix\doi{10.1007/978-3-319-93935-3}.
%Type = Article
\bibitem[{Belkin(1980)}]{belkin1980anomalous}
\bibinfo{author}{Belkin, N.J.}, \bibinfo{year}{1980}.
\newblock \bibinfo{title}{Anomalous states of knowledge as a basis for
  information retrieval}.
\newblock \bibinfo{journal}{Canadian Journal of Information Science}
  \bibinfo{volume}{5}, \bibinfo{pages}{133--143}.
%Type = Article
\bibitem[{Bizer et~al.(2009)Bizer, Lehmann, Kobilarov, Auer, Becker, Cyganiak
  and Hellmann}]{dbpedia}
\bibinfo{author}{Bizer, C.}, \bibinfo{author}{Lehmann, J.},
  \bibinfo{author}{Kobilarov, G.}, \bibinfo{author}{Auer, S.},
  \bibinfo{author}{Becker, C.}, \bibinfo{author}{Cyganiak, R.},
  \bibinfo{author}{Hellmann, S.}, \bibinfo{year}{2009}.
\newblock \bibinfo{title}{Dbpedia - a crystallization point for the web of
  data}.
\newblock \bibinfo{journal}{Web Semant.} \bibinfo{volume}{7},
  \bibinfo{pages}{154–165}.
\newblock \URLprefix \url{https://doi.org/10.1016/j.websem.2009.07.002},
  \DOIprefix\doi{10.1016/j.websem.2009.07.002}.
%Type = Article
\bibitem[{Croft and Thompson(1987)}]{croft_newapproach_i3r}
\bibinfo{author}{Croft, W.B.}, \bibinfo{author}{Thompson, R.H.},
  \bibinfo{year}{1987}.
\newblock \bibinfo{title}{I3r: A new approach to the design of document
  retrieval systems}.
\newblock \bibinfo{journal}{JASIST} \bibinfo{volume}{38},
  \bibinfo{pages}{389--404}.
%Type = Inproceedings
\bibitem[{Daiber et~al.(2013)Daiber, Jakob, Hokamp and
  Mendes}]{dbpediaspotlight}
\bibinfo{author}{Daiber, J.}, \bibinfo{author}{Jakob, M.},
  \bibinfo{author}{Hokamp, C.}, \bibinfo{author}{Mendes, P.N.},
  \bibinfo{year}{2013}.
\newblock \bibinfo{title}{Improving efficiency and accuracy in multilingual
  entity extraction}, in: \bibinfo{booktitle}{Proceedings of the 9th
  International Conference on Semantic Systems (I-Semantics)}.
%Type = Inproceedings
\bibitem[{Dalton et~al.(2014)Dalton, Dietz and Allan}]{entity_query_expansion}
\bibinfo{author}{Dalton, J.}, \bibinfo{author}{Dietz, L.},
  \bibinfo{author}{Allan, J.}, \bibinfo{year}{2014}.
\newblock \bibinfo{title}{Entity query feature expansion using knowledge base
  links}, in: \bibinfo{booktitle}{Proceedings of the 37th International ACM
  SIGIR Conference on Research \& Development in Information Retrieval},
  \bibinfo{publisher}{Association for Computing Machinery},
  \bibinfo{address}{New York, NY, USA}. p. \bibinfo{pages}{365–374}.
\newblock \URLprefix \url{https://doi.org/10.1145/2600428.2609628},
  \DOIprefix\doi{10.1145/2600428.2609628}.
%Type = Article
\bibitem[{Dalton et~al.(2020a)Dalton, Xiong and Callan}]{castoverview}
\bibinfo{author}{Dalton, J.}, \bibinfo{author}{Xiong, C.},
  \bibinfo{author}{Callan, J.}, \bibinfo{year}{2020}a.
\newblock \bibinfo{title}{{TREC} cast 2019: The conversational assistance track
  overview}.
\newblock \bibinfo{journal}{CoRR} \bibinfo{volume}{abs/2003.13624}.
\newblock \URLprefix \url{https://arxiv.org/abs/2003.13624},
  \href{http://arxiv.org/abs/2003.13624}{\tt arXiv:2003.13624}.
%Type = Misc
\bibitem[{Dalton et~al.(2020b)Dalton, Xiong and Callan}]{trecCast}
\bibinfo{author}{Dalton, J.}, \bibinfo{author}{Xiong, C.},
  \bibinfo{author}{Callan, J.}, \bibinfo{year}{2020}b.
\newblock \bibinfo{title}{The trec conversational assistance track (cast)}.
\newblock \URLprefix \url{http://www.treccast.ai/}.
%Type = Article
\bibitem[{Devlin et~al.(2018)Devlin, Chang, Lee and Toutanova}]{bertOriginal}
\bibinfo{author}{Devlin, J.}, \bibinfo{author}{Chang, M.},
  \bibinfo{author}{Lee, K.}, \bibinfo{author}{Toutanova, K.},
  \bibinfo{year}{2018}.
\newblock \bibinfo{title}{{BERT:} pre-training of deep bidirectional
  transformers for language understanding}.
\newblock \bibinfo{journal}{CoRR} \bibinfo{volume}{abs/1810.04805}.
\newblock \URLprefix \url{http://arxiv.org/abs/1810.04805},
  \href{http://arxiv.org/abs/1810.04805}{\tt arXiv:1810.04805}.
%Type = Misc
\bibitem[{Dietz et~al.(2018)Dietz, Gamari and Dalton}]{treccar}
\bibinfo{author}{Dietz, L.}, \bibinfo{author}{Gamari, B.},
  \bibinfo{author}{Dalton, J.}, \bibinfo{year}{2018}.
\newblock \bibinfo{title}{Trec car 2.1: A data set for complex answer
  retrieval}.
\newblock \URLprefix \url{http://trec-car.cs.unh.edu}.
%Type = Article
\bibitem[{Dinan et~al.(2018)Dinan, Roller, Shuster, Fan, Auli and
  Weston}]{wizardofwikipedia}
\bibinfo{author}{Dinan, E.}, \bibinfo{author}{Roller, S.},
  \bibinfo{author}{Shuster, K.}, \bibinfo{author}{Fan, A.},
  \bibinfo{author}{Auli, M.}, \bibinfo{author}{Weston, J.},
  \bibinfo{year}{2018}.
\newblock \bibinfo{title}{Wizard of wikipedia: Knowledge-powered conversational
  agents}.
\newblock \bibinfo{journal}{CoRR} \bibinfo{volume}{abs/1811.01241}.
\newblock \URLprefix \url{http://arxiv.org/abs/1811.01241},
  \href{http://arxiv.org/abs/1811.01241}{\tt arXiv:1811.01241}.
%Type = Inproceedings
\bibitem[{Elgohary et~al.(2019)Elgohary, Peskov and
  Boyd{-}Graber}]{canYouUnpackIt}
\bibinfo{author}{Elgohary, A.}, \bibinfo{author}{Peskov, D.},
  \bibinfo{author}{Boyd{-}Graber, J.L.}, \bibinfo{year}{2019}.
\newblock \bibinfo{title}{Can you unpack that? learning to rewrite
  questions-in-context}, in: \bibinfo{editor}{Inui, K.},
  \bibinfo{editor}{Jiang, J.}, \bibinfo{editor}{Ng, V.}, \bibinfo{editor}{Wan,
  X.} (Eds.), \bibinfo{booktitle}{Proceedings of the 2019 Conference on
  Empirical Methods in Natural Language Processing and the 9th International
  Joint Conference on Natural Language Processing, {EMNLP-IJCNLP} 2019, Hong
  Kong, China, November 3-7, 2019}, \bibinfo{publisher}{Association for
  Computational Linguistics}. pp. \bibinfo{pages}{5917--5923}.
\newblock \URLprefix \url{https://doi.org/10.18653/v1/D19-1605},
  \DOIprefix\doi{10.18653/v1/D19-1605}.
%Type = Article
\bibitem[{Han et~al.(2020)Han, Wang, Bendersky and
  Najork}]{han2020learningtorank}
\bibinfo{author}{Han, S.}, \bibinfo{author}{Wang, X.},
  \bibinfo{author}{Bendersky, M.}, \bibinfo{author}{Najork, M.},
  \bibinfo{year}{2020}.
\newblock \bibinfo{title}{Learning-to-rank with {BERT} in tf-ranking}.
\newblock \bibinfo{journal}{CoRR} \bibinfo{volume}{abs/2004.08476}.
\newblock \URLprefix \url{https://arxiv.org/abs/2004.08476},
  \href{http://arxiv.org/abs/2004.08476}{\tt arXiv:2004.08476}.
%Type = Inproceedings
\bibitem[{Hermann et~al.(2015)Hermann, Kocisky, Grefenstette, Espeholt, Kay,
  Suleyman and Blunsom}]{cnndailymail}
\bibinfo{author}{Hermann, K.M.}, \bibinfo{author}{Kocisky, T.},
  \bibinfo{author}{Grefenstette, E.}, \bibinfo{author}{Espeholt, L.},
  \bibinfo{author}{Kay, W.}, \bibinfo{author}{Suleyman, M.},
  \bibinfo{author}{Blunsom, P.}, \bibinfo{year}{2015}.
\newblock \bibinfo{title}{Teaching machines to read and comprehend}, in:
  \bibinfo{booktitle}{Advances in neural information processing systems}, pp.
  \bibinfo{pages}{1693--1701}.
%Type = Inproceedings
\bibitem[{Hoffart et~al.(2011)Hoffart, Yosef, Bordino, F{\"u}rstenau, Pinkal,
  Spaniol, Taneva, Thater and Weikum}]{ELaida}
\bibinfo{author}{Hoffart, J.}, \bibinfo{author}{Yosef, M.A.},
  \bibinfo{author}{Bordino, I.}, \bibinfo{author}{F{\"u}rstenau, H.},
  \bibinfo{author}{Pinkal, M.}, \bibinfo{author}{Spaniol, M.},
  \bibinfo{author}{Taneva, B.}, \bibinfo{author}{Thater, S.},
  \bibinfo{author}{Weikum, G.}, \bibinfo{year}{2011}.
\newblock \bibinfo{title}{Robust disambiguation of named entities in text}, in:
  \bibinfo{booktitle}{Proceedings of the 2011 Conference on Empirical Methods
  in Natural Language Processing}, \bibinfo{publisher}{Association for
  Computational Linguistics}, \bibinfo{address}{Edinburgh, Scotland, UK.}. pp.
  \bibinfo{pages}{782--792}.
\newblock \URLprefix \url{https://www.aclweb.org/anthology/D11-1072}.
%Type = Article
\bibitem[{Huang et~al.(2020)Huang, Zhu and Gao}]{challenges_dialog_systems}
\bibinfo{author}{Huang, M.}, \bibinfo{author}{Zhu, X.}, \bibinfo{author}{Gao,
  J.}, \bibinfo{year}{2020}.
\newblock \bibinfo{title}{Challenges in building intelligent open-domain dialog
  systems}.
\newblock \bibinfo{journal}{{ACM} Trans. Inf. Syst.} \bibinfo{volume}{38},
  \bibinfo{pages}{21:1--21:32}.
\newblock \URLprefix \url{https://doi.org/10.1145/3383123},
  \DOIprefix\doi{10.1145/3383123}.
%Type = Inproceedings
\bibitem[{Ishigaki et~al.(2020)Ishigaki, Huang, Takamura, Chen and
  Okumura}]{query_biased_abstractive_summarization}
\bibinfo{author}{Ishigaki, T.}, \bibinfo{author}{Huang, H.H.},
  \bibinfo{author}{Takamura, H.}, \bibinfo{author}{Chen, H.H.},
  \bibinfo{author}{Okumura, M.}, \bibinfo{year}{2020}.
\newblock \bibinfo{title}{Neural query-biased abstractive summarization using
  copying mechanism}, in: \bibinfo{editor}{Jose, J.M.},
  \bibinfo{editor}{Yilmaz, E.}, \bibinfo{editor}{Magalh{\~a}es, J.},
  \bibinfo{editor}{Castells, P.}, \bibinfo{editor}{Ferro, N.},
  \bibinfo{editor}{Silva, M.J.}, \bibinfo{editor}{Martins, F.} (Eds.),
  \bibinfo{booktitle}{Advances in Information Retrieval},
  \bibinfo{publisher}{Springer International Publishing},
  \bibinfo{address}{Cham}. pp. \bibinfo{pages}{174--181}.
%Type = Inproceedings
\bibitem[{Kato et~al.(2020)Kato, Imrattanatrai, Yamamoto, Ohshima and
  Tanaka}]{context_ranking_entities}
\bibinfo{author}{Kato, M.P.}, \bibinfo{author}{Imrattanatrai, W.},
  \bibinfo{author}{Yamamoto, T.}, \bibinfo{author}{Ohshima, H.},
  \bibinfo{author}{Tanaka, K.}, \bibinfo{year}{2020}.
\newblock \bibinfo{title}{Context-guided learning to rank entities}, in:
  \bibinfo{editor}{Jose, J.M.}, \bibinfo{editor}{Yilmaz, E.},
  \bibinfo{editor}{Magalh{\~a}es, J.}, \bibinfo{editor}{Castells, P.},
  \bibinfo{editor}{Ferro, N.}, \bibinfo{editor}{Silva, M.J.},
  \bibinfo{editor}{Martins, F.} (Eds.), \bibinfo{booktitle}{Advances in
  Information Retrieval}, \bibinfo{publisher}{Springer International
  Publishing}, \bibinfo{address}{Cham}. pp. \bibinfo{pages}{83--96}.
%Type = Inproceedings
\bibitem[{Li et~al.(2016)Li, Monroe, Ritter, Jurafsky, Galley and
  Gao}]{reinforcement_dialog_generation}
\bibinfo{author}{Li, J.}, \bibinfo{author}{Monroe, W.},
  \bibinfo{author}{Ritter, A.}, \bibinfo{author}{Jurafsky, D.},
  \bibinfo{author}{Galley, M.}, \bibinfo{author}{Gao, J.},
  \bibinfo{year}{2016}.
\newblock \bibinfo{title}{Deep reinforcement learning for dialogue generation},
  in: \bibinfo{booktitle}{Proceedings of the 2016 Conference on Empirical
  Methods in Natural Language Processing}, \bibinfo{publisher}{Association for
  Computational Linguistics}, \bibinfo{address}{Austin, Texas}. pp.
  \bibinfo{pages}{1192--1202}.
\newblock \URLprefix \url{https://www.aclweb.org/anthology/D16-1127},
  \DOIprefix\doi{10.18653/v1/D16-1127}.
%Type = Inproceedings
\bibitem[{Li et~al.(2017)Li, Monroe, Shi, Jean, Ritter and
  Jurafsky}]{adversarial_dialogue_generation}
\bibinfo{author}{Li, J.}, \bibinfo{author}{Monroe, W.}, \bibinfo{author}{Shi,
  T.}, \bibinfo{author}{Jean, S.}, \bibinfo{author}{Ritter, A.},
  \bibinfo{author}{Jurafsky, D.}, \bibinfo{year}{2017}.
\newblock \bibinfo{title}{Adversarial learning for neural dialogue generation},
  in: \bibinfo{booktitle}{Proceedings of the 2017 Conference on Empirical
  Methods in Natural Language Processing}, \bibinfo{publisher}{Association for
  Computational Linguistics}, \bibinfo{address}{Copenhagen, Denmark}. pp.
  \bibinfo{pages}{2157--2169}.
\newblock \URLprefix \url{https://www.aclweb.org/anthology/D17-1230},
  \DOIprefix\doi{10.18653/v1/D17-1230}.
%Type = Article
\bibitem[{Lin et~al.(2020)Lin, Yang, Nogueira, Tsai, Wang and
  Lin}]{t5conversational}
\bibinfo{author}{Lin, S.}, \bibinfo{author}{Yang, J.},
  \bibinfo{author}{Nogueira, R.}, \bibinfo{author}{Tsai, M.},
  \bibinfo{author}{Wang, C.}, \bibinfo{author}{Lin, J.}, \bibinfo{year}{2020}.
\newblock \bibinfo{title}{Conversational question reformulation via
  sequence-to-sequence architectures and pretrained language models}.
\newblock \bibinfo{journal}{CoRR} \bibinfo{volume}{abs/2004.01909}.
\newblock \URLprefix \url{https://arxiv.org/abs/2004.01909},
  \href{http://arxiv.org/abs/2004.01909}{\tt arXiv:2004.01909}.
%Type = Article
\bibitem[{Liu et~al.(2019)Liu, Ott, Goyal, Du, Joshi, Chen, Levy, Lewis,
  Zettlemoyer and Stoyanov}]{roberta}
\bibinfo{author}{Liu, Y.}, \bibinfo{author}{Ott, M.}, \bibinfo{author}{Goyal,
  N.}, \bibinfo{author}{Du, J.}, \bibinfo{author}{Joshi, M.},
  \bibinfo{author}{Chen, D.}, \bibinfo{author}{Levy, O.},
  \bibinfo{author}{Lewis, M.}, \bibinfo{author}{Zettlemoyer, L.},
  \bibinfo{author}{Stoyanov, V.}, \bibinfo{year}{2019}.
\newblock \bibinfo{title}{Roberta: {A} robustly optimized {BERT} pretraining
  approach}.
\newblock \bibinfo{journal}{CoRR} \bibinfo{volume}{abs/1907.11692}.
\newblock \URLprefix \url{http://arxiv.org/abs/1907.11692},
  \href{http://arxiv.org/abs/1907.11692}{\tt arXiv:1907.11692}.
%Type = Inproceedings
\bibitem[{Milne and Witten(2008)}]{relatednessMeasure}
\bibinfo{author}{Milne, D.}, \bibinfo{author}{Witten, I.H.},
  \bibinfo{year}{2008}.
\newblock \bibinfo{title}{Learning to link with wikipedia}, in:
  \bibinfo{booktitle}{Proceedings of the 17th ACM Conference on Information and
  Knowledge Management}, \bibinfo{publisher}{Association for Computing
  Machinery}, \bibinfo{address}{New York, NY, USA}. p.
  \bibinfo{pages}{509–518}.
\newblock \URLprefix \url{https://doi.org/10.1145/1458082.1458150},
  \DOIprefix\doi{10.1145/1458082.1458150}.
%Type = Article
\bibitem[{Nguyen et~al.(2016)Nguyen, Rosenberg, Song, Gao, Tiwary, Majumder and
  Deng}]{marcoDataset}
\bibinfo{author}{Nguyen, T.}, \bibinfo{author}{Rosenberg, M.},
  \bibinfo{author}{Song, X.}, \bibinfo{author}{Gao, J.},
  \bibinfo{author}{Tiwary, S.}, \bibinfo{author}{Majumder, R.},
  \bibinfo{author}{Deng, L.}, \bibinfo{year}{2016}.
\newblock \bibinfo{title}{{MS} {MARCO:} {A} human generated machine reading
  comprehension dataset}.
\newblock \bibinfo{journal}{CoRR} \bibinfo{volume}{abs/1611.09268}.
\newblock \URLprefix \url{http://arxiv.org/abs/1611.09268},
  \href{http://arxiv.org/abs/1611.09268}{\tt arXiv:1611.09268}.
%Type = Misc
\bibitem[{NIST(2019)}]{washingtonDataset}
\bibinfo{author}{NIST}, \bibinfo{year}{2019}.
\newblock \bibinfo{title}{Trec washington post corpus}.
\newblock \URLprefix \url{https://trec.nist.gov/data/wapost/}.
%Type = Article
\bibitem[{Nogueira and Cho(2019)}]{passagererankingbert}
\bibinfo{author}{Nogueira, R.}, \bibinfo{author}{Cho, K.},
  \bibinfo{year}{2019}.
\newblock \bibinfo{title}{Passage re-ranking with {BERT}}.
\newblock \bibinfo{journal}{CoRR} \bibinfo{volume}{abs/1901.04085}.
\newblock \URLprefix \url{http://arxiv.org/abs/1901.04085},
  \href{http://arxiv.org/abs/1901.04085}{\tt arXiv:1901.04085}.
%Type = Article
\bibitem[{Nogueira et~al.(2019)Nogueira, Yang, Cho and
  Lin}]{nogueira2019multistage}
\bibinfo{author}{Nogueira, R.}, \bibinfo{author}{Yang, W.},
  \bibinfo{author}{Cho, K.}, \bibinfo{author}{Lin, J.}, \bibinfo{year}{2019}.
\newblock \bibinfo{title}{Multi-stage document ranking with {BERT}}.
\newblock \bibinfo{journal}{CoRR} \bibinfo{volume}{abs/1910.14424}.
\newblock \URLprefix \url{http://arxiv.org/abs/1910.14424},
  \href{http://arxiv.org/abs/1910.14424}{\tt arXiv:1910.14424}.
%Type = Article
\bibitem[{Oddy(1977)}]{oddy_information_1977}
\bibinfo{author}{Oddy, R.N.}, \bibinfo{year}{1977}.
\newblock \bibinfo{title}{Information retrieval through man-machine dialogue}.
\newblock \bibinfo{journal}{Journal of Documentation} \bibinfo{volume}{33},
  \bibinfo{pages}{1--14}.
%Type = Inproceedings
\bibitem[{Piccinno and Ferragina(2014)}]{ELwat}
\bibinfo{author}{Piccinno, F.}, \bibinfo{author}{Ferragina, P.},
  \bibinfo{year}{2014}.
\newblock \bibinfo{title}{From tagme to wat: A new entity annotator}, in:
  \bibinfo{booktitle}{Proceedings of the First International Workshop on Entity
  Recognition \&; Disambiguation}, \bibinfo{publisher}{Association for
  Computing Machinery}, \bibinfo{address}{New York, NY, USA}. p.
  \bibinfo{pages}{55–62}.
\newblock \URLprefix \url{https://doi.org/10.1145/2633211.2634350},
  \DOIprefix\doi{10.1145/2633211.2634350}.
%Type = Inproceedings
\bibitem[{Qin et~al.(2019)Qin, Liu, Che, Wen, Li and
  Liu}]{qin-etal-2019-entity}
\bibinfo{author}{Qin, L.}, \bibinfo{author}{Liu, Y.}, \bibinfo{author}{Che,
  W.}, \bibinfo{author}{Wen, H.}, \bibinfo{author}{Li, Y.},
  \bibinfo{author}{Liu, T.}, \bibinfo{year}{2019}.
\newblock \bibinfo{title}{Entity-consistent end-to-end task-oriented dialogue
  system with {KB} retriever}, in: \bibinfo{booktitle}{Proceedings of the 2019
  Conference on Empirical Methods in Natural Language Processing and the 9th
  International Joint Conference on Natural Language Processing
  (EMNLP-IJCNLP)}, \bibinfo{publisher}{Association for Computational
  Linguistics}, \bibinfo{address}{Hong Kong, China}. pp.
  \bibinfo{pages}{133--142}.
\newblock \URLprefix \url{https://www.aclweb.org/anthology/D19-1013},
  \DOIprefix\doi{10.18653/v1/D19-1013}.
%Type = Inproceedings
\bibitem[{Qu et~al.(2020)Qu, Yang, Chen, Qiu, Croft and
  Iyyer}]{open-retrieval-qa-sigir2020}
\bibinfo{author}{Qu, C.}, \bibinfo{author}{Yang, L.}, \bibinfo{author}{Chen,
  C.}, \bibinfo{author}{Qiu, M.}, \bibinfo{author}{Croft, W.B.},
  \bibinfo{author}{Iyyer, M.}, \bibinfo{year}{2020}.
\newblock \bibinfo{title}{Open-retrieval conversational question answering},
  in: \bibinfo{booktitle}{Proceedings of the 43rd International ACM SIGIR
  Conference on Research and Development in Information Retrieval},
  \bibinfo{publisher}{Association for Computing Machinery},
  \bibinfo{address}{New York, NY, USA}. p. \bibinfo{pages}{539–548}.
\newblock \URLprefix \url{https://doi.org/10.1145/3397271.3401110},
  \DOIprefix\doi{10.1145/3397271.3401110}.
%Type = Article
\bibitem[{Raffel et~al.(2019)Raffel, Shazeer, Roberts, Lee, Narang, Matena,
  Zhou, Li and Liu}]{t5}
\bibinfo{author}{Raffel, C.}, \bibinfo{author}{Shazeer, N.},
  \bibinfo{author}{Roberts, A.}, \bibinfo{author}{Lee, K.},
  \bibinfo{author}{Narang, S.}, \bibinfo{author}{Matena, M.},
  \bibinfo{author}{Zhou, Y.}, \bibinfo{author}{Li, W.}, \bibinfo{author}{Liu,
  P.J.}, \bibinfo{year}{2019}.
\newblock \bibinfo{title}{Exploring the limits of transfer learning with a
  unified text-to-text transformer}.
\newblock \bibinfo{journal}{CoRR} \bibinfo{volume}{abs/1910.10683}.
\newblock \URLprefix \url{http://arxiv.org/abs/1910.10683},
  \href{http://arxiv.org/abs/1910.10683}{\tt arXiv:1910.10683}.
%Type = Inproceedings
\bibitem[{Song et~al.(2018)Song, Li, Nie, Zhang, Zhao and
  Yan}]{retrieval_based_generation_1}
\bibinfo{author}{Song, Y.}, \bibinfo{author}{Li, C.T.}, \bibinfo{author}{Nie,
  J.Y.}, \bibinfo{author}{Zhang, M.}, \bibinfo{author}{Zhao, D.},
  \bibinfo{author}{Yan, R.}, \bibinfo{year}{2018}.
\newblock \bibinfo{title}{An ensemble of retrieval-based and generation-based
  human-computer conversation systems}, in: \bibinfo{booktitle}{Proceedings of
  the Twenty-Seventh International Joint Conference on Artificial Intelligence,
  {IJCAI-18}}, \bibinfo{publisher}{International Joint Conferences on
  Artificial Intelligence Organization}. pp. \bibinfo{pages}{4382--4388}.
\newblock \URLprefix \url{https://doi.org/10.24963/ijcai.2018/609},
  \DOIprefix\doi{10.24963/ijcai.2018/609}.
%Type = Incollection
\bibitem[{Speck and {Ngonga Ngomo}(2014)}]{fox}
\bibinfo{author}{Speck, R.}, \bibinfo{author}{{Ngonga Ngomo}, A.C.},
  \bibinfo{year}{2014}.
\newblock \bibinfo{title}{Ensemble learning for named entity recognition}, in:
  \bibinfo{booktitle}{The Semantic Web – ISWC 2014}.
  \bibinfo{publisher}{Springer International Publishing}. volume
  \bibinfo{volume}{8796} of \textit{\bibinfo{series}{Lecture Notes in Computer
  Science}}.
%Type = Article
\bibitem[{Vaswani et~al.(2017)Vaswani, Shazeer, Parmar, Uszkoreit, Jones,
  Gomez, Kaiser and Polosukhin}]{vaswani2017attention}
\bibinfo{author}{Vaswani, A.}, \bibinfo{author}{Shazeer, N.},
  \bibinfo{author}{Parmar, N.}, \bibinfo{author}{Uszkoreit, J.},
  \bibinfo{author}{Jones, L.}, \bibinfo{author}{Gomez, A.N.},
  \bibinfo{author}{Kaiser, L.}, \bibinfo{author}{Polosukhin, I.},
  \bibinfo{year}{2017}.
\newblock \bibinfo{title}{Attention is all you need}.
\newblock \bibinfo{journal}{CoRR} \bibinfo{volume}{abs/1706.03762}.
\newblock \URLprefix \url{http://arxiv.org/abs/1706.03762},
  \href{http://arxiv.org/abs/1706.03762}{\tt arXiv:1706.03762}.
%Type = Article
\bibitem[{Voskarides et~al.(2020)Voskarides, Li, Ren, Kanoulas and
  de~Rijke}]{limited_supervision_query_rewrite}
\bibinfo{author}{Voskarides, N.}, \bibinfo{author}{Li, D.},
  \bibinfo{author}{Ren, P.}, \bibinfo{author}{Kanoulas, E.},
  \bibinfo{author}{de~Rijke, M.}, \bibinfo{year}{2020}.
\newblock \bibinfo{title}{Query resolution for conversational search with
  limited supervision}.
\newblock \bibinfo{journal}{Proceedings of the 43rd International ACM SIGIR
  Conference on Research and Development in Information Retrieval} \URLprefix
  \url{http://dx.doi.org/10.1145/3397271.3401130},
  \DOIprefix\doi{10.1145/3397271.3401130}.
%Type = Inproceedings
\bibitem[{Vtyurina et~al.(2017)Vtyurina, Savenkov, Agichtein and
  Clarke}]{evaluation_conversational_assistants}
\bibinfo{author}{Vtyurina, A.}, \bibinfo{author}{Savenkov, D.},
  \bibinfo{author}{Agichtein, E.}, \bibinfo{author}{Clarke, C.L.A.},
  \bibinfo{year}{2017}.
\newblock \bibinfo{title}{Exploring conversational search with humans,
  assistants, and wizards}, in: \bibinfo{booktitle}{Proceedings of the 2017 CHI
  Conference Extended Abstracts on Human Factors in Computing Systems},
  \bibinfo{publisher}{Association for Computing Machinery},
  \bibinfo{address}{New York, NY, USA}. p. \bibinfo{pages}{2187–2193}.
\newblock \URLprefix \url{https://doi.org/10.1145/3027063.3053175},
  \DOIprefix\doi{10.1145/3027063.3053175}.
%Type = Article
\bibitem[{Wang et~al.(2020)Wang, Liu, Bi, Liu, He, Xu and
  Yang}]{knowledge-aware-dialogue-generation}
\bibinfo{author}{Wang, J.}, \bibinfo{author}{Liu, J.}, \bibinfo{author}{Bi,
  W.}, \bibinfo{author}{Liu, X.}, \bibinfo{author}{He, K.},
  \bibinfo{author}{Xu, R.}, \bibinfo{author}{Yang, M.}, \bibinfo{year}{2020}.
\newblock \bibinfo{title}{Improving knowledge-aware dialogue generation via
  knowledge base question answering}.
\newblock \bibinfo{journal}{Proceedings of the AAAI Conference on Artificial
  Intelligence} \bibinfo{volume}{34}, \bibinfo{pages}{9169--9176}.
\newblock \DOIprefix\doi{10.1609/aaai.v34i05.6453}.
%Type = Inproceedings
\bibitem[{Xiong et~al.(2018)Xiong, Liu, Callan and
  Liu}]{entity_saliency_retrieval}
\bibinfo{author}{Xiong, C.}, \bibinfo{author}{Liu, Z.},
  \bibinfo{author}{Callan, J.}, \bibinfo{author}{Liu, T.Y.},
  \bibinfo{year}{2018}.
\newblock \bibinfo{title}{Towards better text understanding and retrieval
  through kernel entity salience modeling}, in: \bibinfo{booktitle}{The 41st
  International ACM SIGIR Conference on Research \&; Development in Information
  Retrieval}, \bibinfo{publisher}{Association for Computing Machinery},
  \bibinfo{address}{New York, NY, USA}. p. \bibinfo{pages}{575–584}.
\newblock \URLprefix \url{https://doi.org/10.1145/3209978.3209982},
  \DOIprefix\doi{10.1145/3209978.3209982}.
%Type = Article
\bibitem[{Yang et~al.(2019)Yang, Dai, Yang, Carbonell, Salakhutdinov and
  Le}]{yang2019xlnet}
\bibinfo{author}{Yang, Z.}, \bibinfo{author}{Dai, Z.}, \bibinfo{author}{Yang,
  Y.}, \bibinfo{author}{Carbonell, J.G.}, \bibinfo{author}{Salakhutdinov, R.},
  \bibinfo{author}{Le, Q.V.}, \bibinfo{year}{2019}.
\newblock \bibinfo{title}{Xlnet: Generalized autoregressive pretraining for
  language understanding}.
\newblock \bibinfo{journal}{CoRR} \bibinfo{volume}{abs/1906.08237}.
\newblock \URLprefix \url{http://arxiv.org/abs/1906.08237},
  \href{http://arxiv.org/abs/1906.08237}{\tt arXiv:1906.08237}.
%Type = Inproceedings
\bibitem[{Zhai and Lafferty(2001)}]{languagemodelsmoothing}
\bibinfo{author}{Zhai, C.}, \bibinfo{author}{Lafferty, J.},
  \bibinfo{year}{2001}.
\newblock \bibinfo{title}{A study of smoothing methods for language models
  applied to ad hoc information retrieval}, in: \bibinfo{booktitle}{Proceedings
  of the 24th Annual International ACM SIGIR Conference on Research and
  Development in Information Retrieval}, \bibinfo{publisher}{Association for
  Computing Machinery}, \bibinfo{address}{New York, NY, USA}. p.
  \bibinfo{pages}{334–342}.
\newblock \URLprefix \url{https://doi.org/10.1145/383952.384019},
  \DOIprefix\doi{10.1145/383952.384019}.
%Type = Inproceedings
\bibitem[{Zhuang et~al.(2017)Zhuang, Wang, Zhang, Xie and
  Zhu}]{retrieval_based_generation_2}
\bibinfo{author}{Zhuang, Y.}, \bibinfo{author}{Wang, X.},
  \bibinfo{author}{Zhang, H.}, \bibinfo{author}{Xie, J.}, \bibinfo{author}{Zhu,
  X.}, \bibinfo{year}{2017}.
\newblock \bibinfo{title}{An ensemble approach to conversation generation}, in:
  \bibinfo{editor}{Huang, X.}, \bibinfo{editor}{Jiang, J.},
  \bibinfo{editor}{Zhao, D.}, \bibinfo{editor}{Feng, Y.},
  \bibinfo{editor}{Hong, Y.} (Eds.), \bibinfo{booktitle}{Natural Language
  Processing and Chinese Computing - 6th {CCF} International Conference,
  {NLPCC} 2017, Dalian, China, November 8-12, 2017, Proceedings},
  \bibinfo{publisher}{Springer}. pp. \bibinfo{pages}{51--62}.
\newblock \URLprefix \url{https://doi.org/10.1007/978-3-319-73618-1\_5},
  \DOIprefix\doi{10.1007/978-3-319-73618-1\_5}.

\end{thebibliography}

\end{document}